\documentclass[11pt]{article}

\usepackage[preprint]{acl}

\makeatletter
\AtBeginDocument{%
  \immediate\write\@auxout{\string\providecommand\string\@LN[2]{}}%
  \immediate\write\@auxout{\string\providecommand\string\@LN@col[1]{}}%
}
\makeatother

\usepackage{times}
\usepackage{latexsym}

\usepackage[T1]{fontenc}

\usepackage[utf8]{inputenc}
\usepackage{textcomp}

\usepackage{microtype}

\IfFileExists{inconsolata.sty}{\usepackage{inconsolata}}{}

\usepackage{graphicx}
\ifdefined\pdfsuppressptexinfo \pdfsuppressptexinfo=-1 \fi

\usepackage{amsmath}

\usepackage{booktabs}
\usepackage{subcaption}
\usepackage{array}
\usepackage{tabularx}       
\usepackage{colortbl}
\usepackage{pgfplots}
\pgfplotsset{compat=1.17}
\usepackage{etoc}
\title{FLUID: Flexible, Evidence-Grounded Situated Task Guidance}

\author{
Yunqi Zhao\thanks{Equal contribution.}\quad
Run Peng\footnotemark[1]\quad
Ze Dong\quad
Ryan Lu\quad
Yinpei Dai\quad
Joyce Chai\\
University of Michigan\\
\small\texttt{\{rickzhao, roihn, chaijy\}@umich.edu}
}

\newcommand{\Ours}{FLUID-Agent}

\newcommand{\roihn}[1]{\textcolor{blue}{\itshape [Roihn: #1]}}
\newcommand{\yunqi}[1]{\textcolor{violet}{\itshape [Yunqi: #1]}}
\newcommand{\ze}[1]{\textcolor{green!50!black}{\itshape [Ze: #1]}}
\newcommand{\jycc}[1]{\textcolor{red}
{\itshape [#1 --JYC]}}
\newcommand{\jyce}[1]{\textcolor{blue}{\itshape [#1 --JYC]}}
\ifdefined\NAACLhidecomments
  \renewcommand{\roihn}[1]{}
  \renewcommand{\yunqi}[1]{}
  \renewcommand{\jycc}[1]{}
  \renewcommand{\jyce}[1]{}
  \renewcommand{\ze}[1]{}
\fi

\newcommand{\pmval}[2]{#1{\scriptsize\,\ensuremath{\pm}#2}}
\newcommand{\pmbest}[2]{\textbf{#1}{\scriptsize\,\ensuremath{\pm}#2}}
\definecolor{tableshade}{HTML}{EDF3FA}

\begin{document}

\maketitle

\begin{abstract}
% Real-time perceptual guidance requires agents to determine not only what assistance to provide, but also why, whether, and when to intervene as a task and interaction unfold. Existing benchmarks typically supervise interventions through response content and timing, but provide limited information about the perceptual and conversational evidence that makes an intervention warranted. We address this gap by augmenting two egocentric interactive datasets with the interaction structure of natural human guidance, including triggering evidence, intervention necessity, and evidence-dependent timing boundaries. Building on this structure, we develop an evidence-grounded evaluation framework that evaluates intervention decisions against their supporting evidence and valid timing. Our evaluation of representative real-time multimodal systems shows that current systems still struggle to recognize when intervention is needed and to identify the evidence that should support it. Inspired by the interaction structure observed in natural human guidance, we develop \Ours{}, a real-time agentic framework that grounds evolving perceptual evidence and reasons about whether, when, and how to intervene. Together, our work establishes human-derived interaction structure as a common foundation for dataset augmentation, evaluation, and system design in real-time perceptual guidance.
Real-time task guidance is inherently flexible: when an agent guides a user through a task, multiple responses and timings can be appropriate, and sometimes the best action is to remain silent. Existing benchmarks often miss this flexibility by treating synthesized dialogues as fixed targets for what to say and when to say it. We introduce \textbf{FLUID} (\textbf{Fl}exible, Evidence-Grounded Situated Task G\textbf{uid}ance), a unified framework for modeling, evaluating, and enabling flexible interaction in multimodal task guidance. Using naturally occurring human-human guidance from WTaG and HoloAssist, FLUID grounds interventions in their supporting evidence, necessity, and timing flexibility, and uses this representation both to evaluate flexible guidance and to build a real-time agent for task guidance. Our agent achieves the strongest overall balance of response correctness, intervention precision, timely guidance, and response latency among evaluated systems, while still leaving room for improvement in avoiding speech overlap. Together, FLUID moves situated task guidance beyond rigid reference matching toward more natural and adaptive human-AI interaction. Project page: \url{https://fluid-guidance.github.io}.
\end{abstract}

%\jycc{(1) fluid situation communication in perceptual guidance has these types of beahviors..... however, most model evaluations based on the overlapping with the ground-truth generation, which often by themsleves are synthezied by LLM. existing metrics do not take interactivity into account. if we want the model to behave like human teachers, we would need to have metrics to evaluate the interactivity. (2) Motivated by natural human guidance, we created the first of its kind evaluation metrics that focus on interactivtiy, (3) we demonstrate that, STOA models, although evaluated competitively  on previous benchmarks, fall sharply on the benchmark for interactivity. need to do some studies e.g., have a pair of system s1 and s2, see their ranking based on previous metrics, vs. their rank based on our metrics, it shows that our ranking are more coorelate with human judgement. So this evaluation metrics will provide a framework for future models that need to engage with humans on task guidance. (4) moreover, based on grounded evidence from human interaction, we incoproate the the design strategies into a n agentic workflow and developed an EgoGuide agent. With expectation, the agentic workflow leads to improved fludity based on our metrics, which is further confirmed by human studies.  }

\section{Introduction}

\begin{figure}[t]
    \centering
    \includegraphics[width=1.0\linewidth]{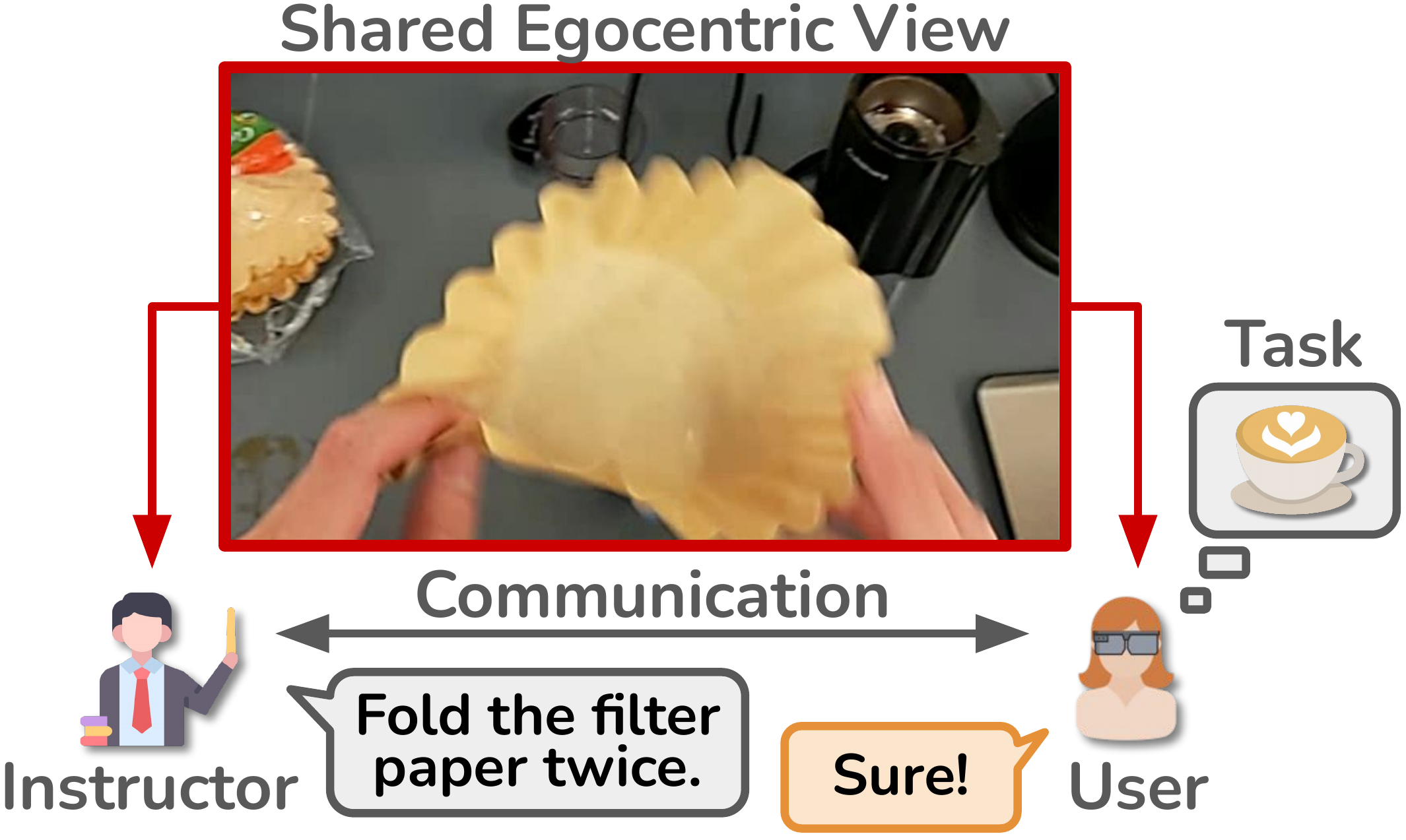}
    \caption{Illustration of perceptual task guidance.}
    \label{fig:illustrative-example}
\end{figure}

%JYC re-organized the intro
With the growing adoption of smart glasses and other wearable devices, multimodal AI assistants are becoming increasingly accessible for providing real-time, situated assistance in everyday activities~\citep{cui2026minicpmo45realtimefullduplex, openai2026gptrealtime2, google2026gemini38live}. An important emerging application is perceptual task guidance, where an AI assistant observes the user's environment and ongoing actions through an egocentric camera and provides context-aware guidance through natural language as shown in Figure~\ref{fig:illustrative-example}. This kind of situated perceptual task guidance requires continuous coordination between visual perception, task understanding, and communication. The AI assistant must interpret the user's actions and task progress, respond to questions, provide step-by-step instructions, detect mistakes, and proactively offer assistance when appropriate. %Crucially, effective guidance requires not only determining what to say, but also when to say it, as communication is tightly intertwined with the evolving task.

% \jycc{a figure to show the interaction}

To evaluate these capabilities, existing work commonly relies on human-human interaction data or synthetically generated dialogues as references for expected guidance behaviors, measuring how closely the AI instructor's responses align with the reference responses in both semantic content and temporal placement~\citep{livecook, zhang2025proactive, shi2026riverrealtimeinteractionbenchmark}. Although a good starting point, these approaches treat reference interactions as fixed targets, overlooking the inherent flexibility of situated guidance. In practice, human-human guidance is fluid and adaptive. Certain situations, such as incorrect actions or imminent safety risks, require immediate intervention, while others allow a wider range of appropriate timing for proactive assistance. Similarly, backchannels and acknowledgments help maintain shared understanding, but their timing and frequency can vary considerably. This creates a fundamental mismatch between rigid reference-based evaluation and the flexible nature of real-world task guidance, calling for evaluation criteria that can take into account both the correctness of guidance and the flexibility of communication.  
%motivating evaluation criteria that adapt to the requirements of each intervention.
%More broadly, existing systems often lack explicit mechanisms for representing and reasoning about such interaction flexibility.

Addressing this challenge requires a principled framework for characterizing the flexibility of situated guidance. Importantly, flexibility does not imply unconstrained behavior. Effective guidance must remain grounded in perceptual and interaction evidence, with different communicative behaviors subject to different requirements. When AI assistants decide to intervene or provide guidance, some interventions (e.g., correcting a user mistake or giving a warning about a potential hazard)  are necessary within narrow temporal windows, while others (e.g., backchannel) are optional and allow greater flexibility in content and timing. Human references remain valuable for identifying effective guidance behaviors, but should not prescribe a single correct response at a fixed moment. Instead, the key is to characterize the underlying evidence that warrants communication and the conditions under which different responses are appropriate. This evidence-grounded perspective provides a foundation for both evaluating and generating flexible guidance behaviors. 

% Layout (page 2): this paragraph is set one line longer, and Figure 1 is at full column width, so that Related Work
% fills the right column exactly; without it the leftover space opens a gap between the Related Work paragraphs.
\looseness=1 In this paper, we introduce FLUID (\textbf{Fl}exible, Evidence-Grounded Situated Task G\textbf{uid}ance), a unified framework for characterizing, evaluating, and enabling flexible interaction dynamics in multimodal task guidance. First, we develop an evidence-grounded representation that characterizes communicative behaviors in terms of their necessity, supporting evidence, and temporal flexibility (\textbf{FLUID-Modeling}). Second, we introduce an evaluation framework that leverages human references while moving beyond strict response matching to assess the appropriateness of guidance content and timing (\textbf{FLUID-Evaluation}). Third, we propose an agent architecture that explicitly models and adapts to evolving interaction dynamics, enabling flexible, evidence-grounded decisions about when and how to communicate, including reactive responses, proactive guidance, and appropriate silence (\textbf{FLUID-Agent}). Together, these contributions provide a systematic approach to understanding, evaluating, and generating flexible, evidence-grounded guidance, advancing multimodal AI assistants toward more natural and adaptive human–AI \mbox{interaction}.

%\jycc{Is there any evaluation/validate to indicate FLUID is a better framework than reference-based evaluation?}

%\jycc{Then the rest of the paper will be organized around the three aspects of FLUID: modeling, evaluation, and agent.  For evaluation, it should define metrics and report the results on these other models, and need to demonstrate the validity of the evaluation. }
\section{Related Work}

% Layout: the space between the paragraphs of this section is fixed. With the style's stretchable skips, any space left
% over in the column (Section 3 cannot start at its foot) is stretched into the gap between the two paragraphs.
\begingroup
\makeatletter
\renewcommand\paragraph{\@startsection{paragraph}{4}{\z@}{1.5ex}{-1em}{\normalsize\bfseries}}
\makeatother
\setlength{\parskip}{0pt}

\paragraph{Real-Time Interaction.}
\looseness=1   % layout (page 2): one line longer, see the note in 01_intro.tex
Recent advances in multimodal systems have rapidly expanded real-time human-AI interaction, from streaming audio-visual perception~\citep{openai2026gptrealtime2,qwenteam2026qwen35omnitechnicalreport} to low-latency, full-duplex spoken interaction~\citep{cui2026minicpmo45realtimefullduplex,thinkingmachines2026interactionmodels,antgroup2026realtimevenusfullduplexinteractionasynchronous,bytedance2026seedrealtime,google2026gemini38live}. Scaling such interactive capabilities has increasingly relied on synthetic or automatically constructed interaction data, often by repurposing existing egocentric videos~\citep{grauman2022ego4dworld3000hours,NEURIPS2024_f4a04396,Damen2018EPICKITCHENS,huang2024egoexolearn} and generating intervention traces at scale~\citep{shi2026riverrealtimeinteractionbenchmark,zhang2025proactive,kundu2026planwatchrecoverbenchmark}. In contrast, naturally occurring human-human interactions in which one person continuously observes and guides another through a physical task remain scarce. Existing efforts include WTaG~\citep{bao2023can}, HoloAssist~\citep{HoloAssist2023}, EgoInstruct~\citep{sakai2025egoinstruct}, Ego-EXTRA~\citep{ragusa2026egoextra}, and Ego-MC-Bench~\citep{bhattacharyya2026streaming}, but some of them remain limited in released modalities. In this work, we primarily build on WTaG and HoloAssist, both of which provide audio alongside egocentric video, to study how interactions naturally unfold during human guidance.

\paragraph{Proactive Agent Modeling.}
Existing proactive video agents span query-conditioned event response, continuous commentary, and situated procedural guidance \citep{wang2025proactivevideoqacomprehensivebenchmarkevaluating, lu2026omniinteract, VideoLLM-online, fu2025vispeak, qian2025dispider, wang2026mmduet, livecc, zhang2025proactive}. Representative systems typically produce textual responses or discrete decisions about whether to intervene, with timing supervision derived from annotated event boundaries, task transitions, mistakes, or synthesized utterances \citep{livefit, zhang2023seagull, zhang2025bootstrapping, livecook, kundu2026planwatchrecoverbenchmark}. Recent methods move beyond purely event-triggered policies: Pro²Assist additionally annotates assistance need, while Vinci2 generates visually grounded service opportunities and reasons about whether intervention is warranted \citep{xu2026pro2assistcontinuousstepawareproactive, gong2026vinci2}. They provide increasingly rich supervision; however, their assistance targets are constructed from task states or annotated events rather than naturally occurring intervention decisions.
% These methods supervise when and what the assistant should say, but do not explicitly link naturally occurring instructor interventions to the local observable evidence that prompted them. We therefore reannotate WTaG and HoloAssist to associate each intervention with its local observable evidence and context-dependent timing window. Because spoken interventions themselves unfold over time, we also require interventions to be realized as speech and jointly evaluate their onset, duration, content, and evidential support.
In contrast, we study human-human guidance.
% and explicitly link each observed instructor intervention to the local perceptual or conversational evidence that elicited it. 
This preserves a broader range of communicative behaviors and enables intervention-specific annotations of necessity and valid timing, characterizing not only when assistance may occur but how much flexibility each interaction permits.
\par
\endgroup

% \paragraph{Efficient Response Generation.}
% \roihn{optional}
% Visual-expert (save redundant tokens (TimeChat), smart caching) keyword: temporal redundancy
% Reasoning in different tracks (Dispider, etc.)

\section{FLUID-Modeling: Grounded Evidence in Task Guidance}
\label{sec:grounded-evidence}
\begin{figure*}[t]
    \centering
    \includegraphics[width=\textwidth]{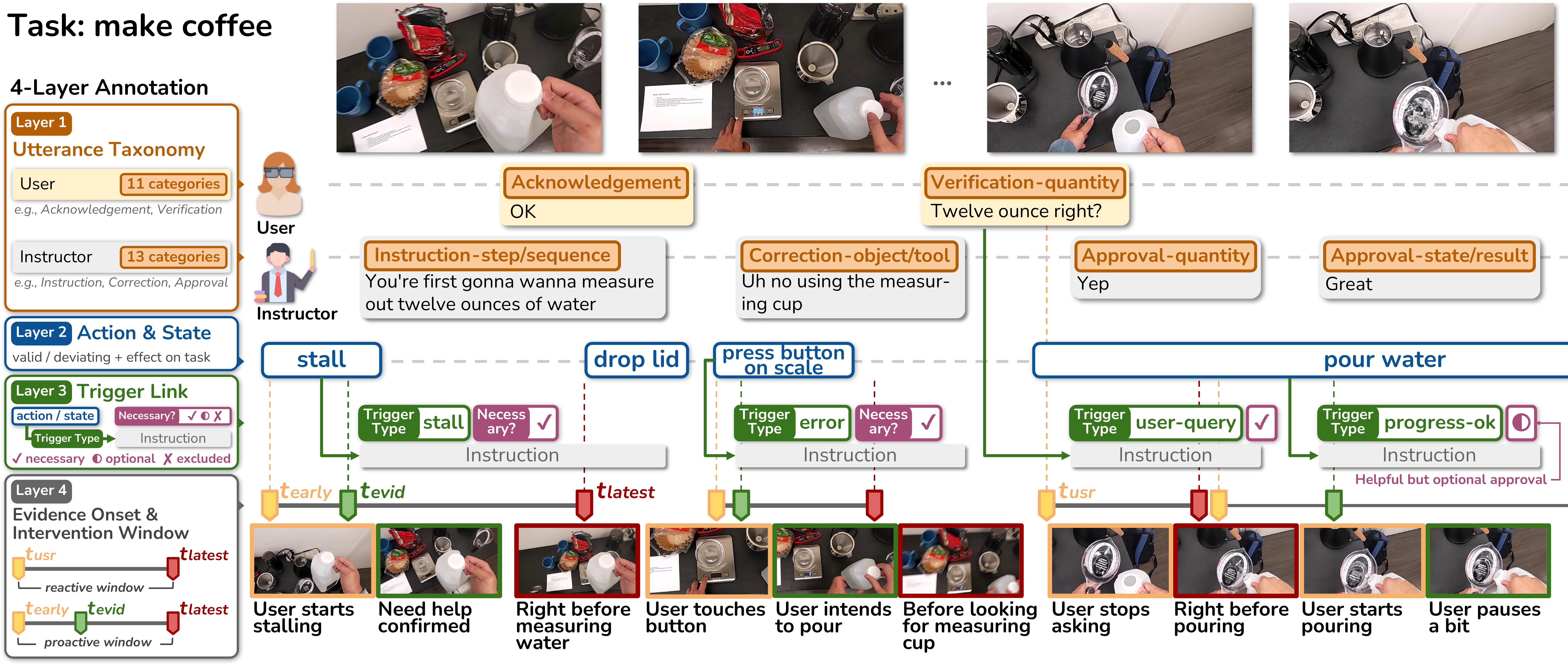}
    \caption{Four-layer annotation of a WTaG coffee session, covering utterance taxonomy, action and state, trigger links, and intervention timing. Each instructor intervention is grounded in its triggering evidence, labeled by necessity, and assigned an evidence-derived intervention window.}
    \label{fig:overview}
\end{figure*}
\subsection{Problem Definition}
\label{sec:problem}

In real-time task guidance, an agent continuously observes a user performing a task and must decide how to respond appropriately as the task and interaction evolve. 
% At every moment, the guide either remains silent or intervenes. 
We characterize each intervention decision along four dimensions. \textbf{WHY}: what evidence warrants the intervention, namely the user utterance, action, or state that the guide reacts to; we refer to it as the \textbf{grounded evidence} of the intervention. \textbf{WHETHER}: whether the intervention is necessary, optional, or not needed. \textbf{WHEN}: the period during which the intervention is useful. \textbf{WHAT}: the interaction act of the intervention and its content. Interventions triggered by a user utterance are reactive; all others are proactive.

\subsection{Source Data}
\label{sec:source-data}
We study WTaG~\citep{bao2023can} and HoloAssist~\citep{HoloAssist2023}, two egocentric datasets of naturally occurring human-human task guidance in which a remote instructor observes the user's view and provides verbal guidance. The two datasets exhibit complementary interaction regimes. WTaG is relatively self-paced: users receive a high-level task recipe in advance and can often proceed independently, which produces mixed-initiative interaction. HoloAssist has a stronger knowledge asymmetry and is predominantly instructor-led, with users relying on ongoing guidance for each step. In total, we select 43 WTaG sessions and 72 HoloAssist sessions with substantive interactions between users and instructors.
% Together, the two datasets let us study intervention decisions across distinct but naturally occurring guidance settings.

% \subsection{Recovering Interaction Structure}
\subsection{Multi-Layer Annotation}
\label{sec:interaction-structure}

To reveal the interaction semantics underlying naturally occurring human interventions, we augment WTaG and HoloAssist with four linked annotation layers, which together cover the four dimensions of Section~\ref{sec:problem} (see Figure~\ref{fig:overview} for an example).

\paragraph{Layer 1: Utterance Taxonomy.}
We categorize every user and instructor utterance using an utterance-level taxonomy with 11 user categories and 13 instructor categories.

\paragraph{Layer 2: Action and State.}
We annotate user actions and states that are closely related to instructor interventions, including hand actions, visual loss, stalling, and searching. Each action is further labeled by whether it is valid or deviating and by its effect on task execution, such as task progress, step completion, or recovery.
%\jycc{why need to emphasize on "closely related to instructor interventions"? I feel this should be state from user activities. even there is no instruction, these states should still be captured? e.g., correct steps don't always get an acknowledgment from the instructor. }
%\roihn{This is a deliberate design choice. There are many possible moments when an instructor could intervene, and exhaustively labeling all of them post hoc would require substantial effort and may not reflect natural ongoing interactions. Instead, we focus only on interventions that actually occurred and investigate what triggered them. In short, we study why interventions happen, rather than annotating every possible moment when an intervention could happen.}

\paragraph{Layer 3: Trigger Link.}
Using the annotated utterances and actions, we link each instructor intervention to the preceding utterance, action, or state that triggered it, forming a \textbf{grounded evidence link} that explains why the human instructor decided to intervene. Each link receives a trigger type derived from the linked events, such as \textit{Step Done}, \textit{Error}, or \textit{User Query}. We further categorize each intervention as \emph{necessary} if, without it, the user would make a mistake, get stuck, or lack needed information; \emph{optional} if the user would proceed correctly on their own, so that speaking and remaining silent are both acceptable; and \emph{excluded} if it is irrelevant to the task, such as self-repairs, or relies on knowledge that cannot be obtained from the shared egocentric view, the dialogue, or the task knowledge.

\paragraph{Layer 4: Evidence Onset and Intervention Window.}
We annotate temporal anchor points for interventions, which define the time window within which an intervention should be initiated. For reactive interventions, we annotate (1) the time at which the user finishes their most recent utterance, $t_{\mathrm{usr}}$, and (2) the latest time at which the intervention remains useful, $t_{\mathrm{latest}}$. For proactive interventions, there is no corresponding $t_{\mathrm{usr}}$; instead, we annotate (1) the earliest time at which initial signs suggest that an intervention may be necessary, $t_{\mathrm{early}}$, and (2) the time at which enough evidence becomes available to determine that the intervention is necessary, $t_{\mathrm{evid}}$, together with $t_{\mathrm{latest}}$. Together, these anchors define an \textbf{evidence-derived window} for evaluating intervention timing.

\paragraph{Annotation Workflow.}
For WTaG, we first use majority voting across multiple language-model annotators to produce initial annotations, followed by careful manual review of all instances. Based on the errors and practical observations identified during this process, we refine the machine annotation pipeline and apply it to the selected HoloAssist sessions, and manually review about 10\% of the generated annotations to ensure the synthetic data quality.
%\jycc{what do you mean experimental release?}
%\roihn{fixed}

\subsection{Interventions Are Context-Dependent}
\label{sec:natural-guidance}

Our annotations show that natural human guidance does not follow a fixed trigger-to-intervention mapping: whether an intervention is needed, and what form it should take, depends on the surrounding interaction context.

\paragraph{Similar situations may or may not require intervention.}
Among step completions on the self-paced WTaG, only 12.9\% are followed by a necessary intervention (e.g., providing the next-step instruction), while 17.5\% are optional (e.g., ``Nice'' or ``uh-huh'') and 69.6\% receive no intervention ($n=418$). In the more instructor-led HoloAssist setting, the corresponding proportions are 65.9\%, 12.2\%, and 21.9\% ($n=649$). Thus, whether a step completion warrants guidance depends strongly on the surrounding interaction regime and context, rather than on the event type alone. In contrast, existing approaches often use a more rigid event-to-response mapping, such as triggering a next-step instruction whenever a step is completed.

%\paragraph{The same trigger can call for different guidance.}
%Interventions following a step completion are mostly approvals on WTaG (43\%, versus 27\% instructions) but mostly instructions on HoloAssist (67\%). Even for the same user intent, the appropriate response depends on the observed state: a verification question may call for approval when the action is valid, but correction when it deviates. Thus, trigger identity alone does not determine what guidance is appropriate; the response must be interpreted against its grounded evidence.

These observations motivate evaluating guidance against the grounded evidence that warrants an intervention, whether intervention is necessary, and when it remains useful. 

\section{FLUID-Evaluation on Task Guidance}
\label{sec:evaluation}
Existing reference-based evaluations assess guidance primarily through content similarity and temporal proximity to human reference utterances, often using a fixed timing window \citep{zhang2025proactive, livecook, bhattacharyya2026streaming}. Such criteria overlook the flexibility of natural interaction, where appropriate guidance depends on task evidence and may vary in necessity and timing. We introduce an evidence-derived evaluation framework that leverages our annotations to evaluate intervention appropriateness based on supporting evidence, necessity, and valid timing windows. Appendix~\ref{app:evaluation} specifies the protocol and metrics in full.

\subsection{Intervention Matching}
\label{sec:eval-matching}
Real-time interaction does not follow fixed turn boundaries: within any time interval, a system may produce zero, one, or multiple utterances, whereas evaluation requires a one-to-one correspondence between system utterances and annotated human interventions. We therefore first match them based on content and supporting evidence. For each necessary or optional intervention, we consider candidate system utterances within a matching window that covers its annotated evidence window and the surrounding intervals of adjacent reference utterances. An LLM judge then assesses content equivalence based on the dialogue history and triggering evidence.
% , assigning a score from 1 (contradiction) to 5 (equivalence). Scores of at least 4 indicate that the user would do or understand the same thing. 
% Each intervention is matched to the earliest qualifying, unmatched system utterance.
Since a system may generate multiple equivalent utterances within the same time interval, we select the earliest qualifying utterance as the match.
%\jycc{this last sentence is confusing. don't know what it means} 
%\roihn{fixed} \jycc{still don't get it, if a system generates multiple equipvalent uttearnces with the same time interval, shouldn't the system be severly punished? cannot simple pick the earliest one and ignore the others. }
%\roihn{Here is about how we match the utterance for further comparison. We have other metrics, like intervention precision, to penalize a system if it generates multiple similar utterances within the same time interval.}

\subsection{Timing Appropriateness}
\label{sec:eval-timing}

Evaluating the timing of matched interventions is essential, yet common approaches ~\citep{zhang2025proactive, livecook, bhattacharyya2026streaming} have two limitations: (1) fixed timing windows ignore when an intervention becomes possible and when it stops being useful, admitting invalid onsets and excluding valid ones; and (2) binary in-window scoring treats all onsets within a window as equally appropriate, although their usefulness varies with timing.

For the first limitation
%\roihn{To me, this paragraph is not for addressing the first limitation. And I do not see where you call back the second limitation in the following paragraphs. This starter also feels like you finish addressing the limitation in this paragraph, while it's not.}\yunqi{fixed}
, we derive intervention-specific valid windows from the temporal anchors annotated in Section~\ref{sec:interaction-structure}: $[t_{\mathrm{usr}},t_{\mathrm{latest}})$ for reactive guidance and $[t_{\mathrm{early}},t_{\mathrm{latest}})$ for proactive guidance. Unlike a fixed timing window around the reference utterance, each window opens after the user's turn ends or at the first sign of need, and closes when the intervention is no longer useful. An onset is valid if it falls anywhere within the window of the corresponding intervention. 

We validate these windows with a human study that isolates timing from content. For each of 79 WTaG instructor utterances, we keep the utterance fixed and sample six candidate onsets around its annotated anchors, spanning early, in-window, and late times and including the instructor's actual onset. Three researchers rate how appropriate it would be to deliver the same utterance at each onset on a 0-10 scale (See Appendix~\ref{app:human-timing-study} for more details). 
We treat an onset as acceptable when its mean human rating is at least 6, and compare this label with whether the onset falls within the proposed intervention window. We report Macro-F1 to measure how well the window classification agrees with human judgments across acceptable and unacceptable timings.
% Treating an onset as acceptable when its mean rating is at least 6, we evaluate each window's ability to distinguish acceptable from unacceptable onsets using Macro-F1, which equally weights classification performance for both classes. 
Overall, our evidence-derived windows outperform the fixed $[-6,+3]$s window around reference utterance onsets adopted from ProAssist~\citep{zhang2025proactive} (Macro-F1: $0.788$ vs.\ $0.713$). The improvement is more pronounced for reactive guidance, where our windows exclude inappropriate onsets that would interrupt the user, while for proactive guidance, our windows achieve comparable performance while admitting earlier evidence-supported onsets missed by the fixed window.

For the second limitation, we introduce a continuous timing score that captures varying degrees of timing appropriateness. The score peaks at a preferred response time $t_{\mathrm{pref}}$ and decreases for earlier or later onsets. For reactive guidance, $t_{\mathrm{pref}}=t_{\mathrm{usr}}$, since responding immediately after the user finishes speaking is preferred; for proactive guidance, $t_{\mathrm{pref}}=t_{\mathrm{evid}}$, when sufficient evidence becomes available. An earlier utterance lacks supporting evidence, and a later one becomes less useful as the task progresses.

Next, we test whether a continuous timing score better reflects human preferences than binary in-window scoring. We start with a simple linear score that decreases from $t_{\mathrm{pref}}$ toward the window boundaries. For each utterance, we compute Spearman's correlation between the scores assigned to the six candidate onsets and their human ratings. Averaged across utterances, the linear score achieves a correlation of $0.83$, compared with $0.65$ for binary in-window scoring.

We further refine the shape of the timing penalty. Let $t$ denote the onset of the system utterance matched to an intervention. For a proactive intervention with $t\in W_E=[t_{\mathrm{early}},t_{\mathrm{evid}})$, we use a quadratic penalty that grows as $t$ moves further before $t_{\mathrm{evid}}$. For $t$ at or after the preferred time, in $W_D=[t_{\mathrm{pref}},t_{\mathrm{latest}})$ for both reactive and proactive interventions, we use a cosine-based decay that penalizes short delays mildly and long delays strongly. We compare this score and the linear score with human ratings rescaled to $[0,1]$ using mean absolute error (MAE). The cosine-based score achieves lower MAE than the linear score (0.182 vs.\ 0.225) and is adopted as the final scoring function:
\begin{equation}
\label{eq:timing-score}
s(t)=
\begin{cases}
1-\lambda_E\left(
\frac{t_{\mathrm{evid}}-t}
{t_{\mathrm{evid}}-t_{\mathrm{early}}}
\right)^2,
& t\in W_E,\\[4pt]
\cos^{p_L}\left(
\frac{\pi(t-t_{\mathrm{pref}})}
{2(t_{\mathrm{latest}}-t_{\mathrm{pref}})}
\right),
& t\in W_D,\\[4pt]
0, & \text{otherwise}.
\end{cases}
\end{equation}

The parameters $\lambda_E$ and $p_L$ control the early penalty and the decay rate, respectively. We set $(\lambda_E,p_L)=(0.7,0.5)$ by cross-validated calibration against the same human ratings, as detailed in Appendix~\ref{app:human-timing-study}.

Based on these, we define \textbf{Timely Correctness} as the average timing score over all necessary interventions. An intervention contributes zero if it has no content-matched system utterance or if that utterance's onset falls outside its window. Optional interventions are excluded, as remaining silent is also acceptable. We additionally report \textbf{Response Latency}, the median delay from $t_{\mathrm{usr}}$ to the start of system speech $t$ for content-matched reactive responses.

\begin{figure*}[t]
    \centering
    \includegraphics[width=\textwidth]{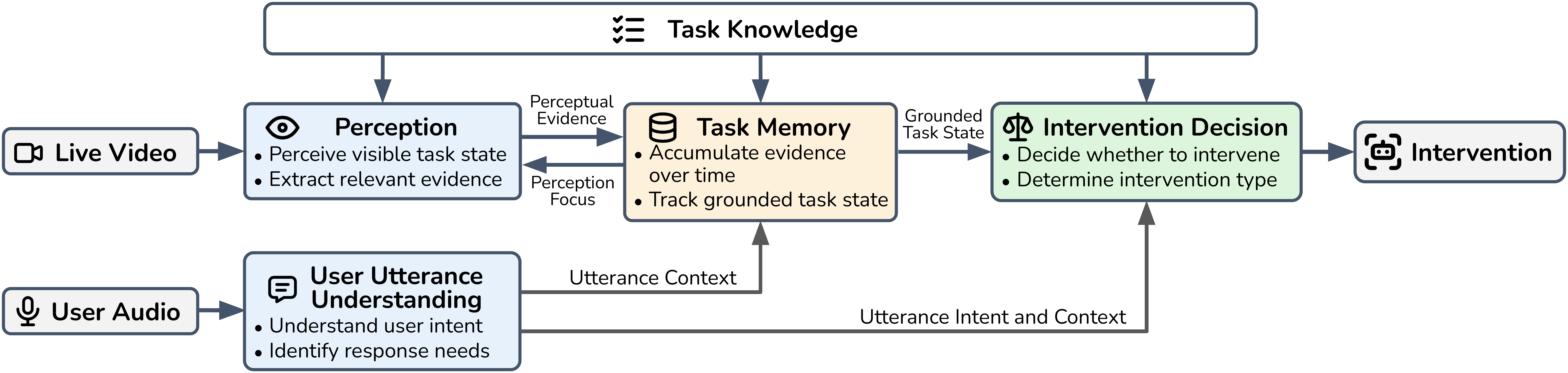}
    \caption{Overview of the \Ours{} framework. Perception, task memory, and user utterance understanding jointly support evidence-grounded intervention decisions as task and interaction context evolve.}
    \label{fig:framework}
\end{figure*}

\subsection{Complementary Evaluation Metrics}
\label{sec:eval-complementary}

\paragraph{Reference Fidelity.}
We measure agreement with human references using two metrics following \citet{zhang2025proactive}. \textbf{Similarity F1} measures embedding-based similarity, while \textbf{LLM-Judged F1} assesses content equivalence using the criterion described above. Both use a fixed timing window around reference utterance onsets for intervention matching.
%\jycc{what is $[-6,+3]$?} 
%\yunqi{fixed}

\paragraph{Intervention Precision and Turn-Taking.}
Timely Correctness measures whether necessary guidance is delivered correctly and on time, but frequent interventions may improve coverage at the cost of unnecessary or disruptive speech. We therefore introduce \textbf{Intervention Precision}, the proportion of all played system utterances that correctly match a necessary or optional intervention within its valid window. Optional interventions are counted as correct when produced, while remaining silent for an optional intervention is not penalized. Unmatched, repeated, incorrect, or mistimed utterances lower this score. Additionally, we report \textbf{Speech Overlap Ratio}, the proportion of user-speaking time overlapping with system speech, to measure potential disruptions to natural turn-taking.

\section{FLUID-Agent: From Interaction Dynamics to Real-Time Guidance}
\label{sec:framework}
Motivated by the interaction dynamics characterized in Section~\ref{sec:grounded-evidence}, \Ours{} is a real-time system for flexible, evidence-grounded task guidance. As shown in Figure~\ref{fig:framework}, it uses an asynchronous agentic architecture built around a single Qwen3.6-35B-A3B vision-language model~\citep{qwen36_35b_a3b}, which serves perception, utterance understanding, and intervention reasoning through task-specific prompting. The components communicate through a shared task memory that tracks task-relevant evidence and interaction context over time (see Appendix~\ref{app:egoguide} for implementation details).

\paragraph{Perception and Task Memory.}
The perception module continuously analyzes keyframes sampled from short temporal windows of the live video stream. Guided by task knowledge and the current task state, it tracks relevant user actions, states, and task progress. These observations, together with user and system utterances and step-level completion status, are maintained in a continuously updated shared task memory, providing grounded task and interaction context for subsequent intervention decisions.

\paragraph{User Utterance Understanding.}
Not every user utterance calls for a response. The user utterance understanding module uses the utterance taxonomy and task context to distinguish utterances that require a response from those that do not, such as task narration, progress reports, and acknowledgments. This enables \Ours{}  to respond when needed while avoiding unnecessary interruptions.

\paragraph{Evidence-Conditioned Intervention Decision.}
Each potential intervention is grounded in evidence from the ongoing task and interaction, such as an observed error, action progress, or user request. The intervention module uses this evidence and the interaction history to reason about whether, when, and how to provide guidance. As the task evolves, \Ours{} reassesses pending interventions, providing evidence-grounded guidance when appropriate or remaining silent when it is no longer warranted.

\paragraph{Asynchronous Execution.}
\Ours{} coordinates perception, utterance understanding, and intervention reasoning asynchronously, allowing each component to process new information as it becomes available. The system continues monitoring the task while speaking, yields to new user requests, and cancels pending guidance when it is no longer relevant.

%\jycc{if need space, this section can be condensed, and move the detail to the appendix. This paper is mostly an evaluation framework paper. }\yunqi{fixed}

\section{Experiments \& Results}
\begin{table*}[t]
\centering

\newlength{\valuewidth}
\begingroup
\large
\setlength{\tabcolsep}{2pt}
\renewcommand{\arraystretch}{1.02}
% Table-local: each value is right-aligned in a box as wide as "0.000", so that
% value ends line up within a column; the CI half-width follows the box.
\settowidth{\valuewidth}{0.000}
\renewcommand{\pmval}[2]{\makebox[\valuewidth][r]{#1}{\scriptsize\,\ensuremath{\pm}#2}}
\renewcommand{\pmbest}[2]{\makebox[\valuewidth][r]{\textbf{#1}}{\scriptsize\,\ensuremath{\pm}#2}}

\resizebox{\textwidth}{!}{%
\begin{tabular}{
    @{}l
    llll
    >{\columncolor{tableshade}}l
    l
    @{\hspace{6pt}}
    llll
    >{\columncolor{tableshade}}l
    l
    @{}
}

\toprule

& \multicolumn{6}{c}{\textbf{WTaG}}
& \multicolumn{6}{c}{\textbf{HoloAssist}} \\

\cmidrule(lr){2-7}
\cmidrule(l){8-13}

\textbf{Metric}
& \multicolumn{1}{c}{\textbf{ProAssist}}
& \multicolumn{1}{c}{\textbf{VideoLLM}}
& \multicolumn{1}{c}{\textbf{JoyAI}}
& \multicolumn{1}{c}{\textbf{Gemini}}
& \multicolumn{1}{c}{\cellcolor{tableshade}\textbf{\begin{tabular}[c]{@{}c@{}}FLUID-\\Agent\end{tabular}}}
& \multicolumn{1}{c}{\textbf{Human}}
& \multicolumn{1}{c}{\textbf{ProAssist}}
& \multicolumn{1}{c}{\textbf{VideoLLM}}
& \multicolumn{1}{c}{\textbf{JoyAI}}
& \multicolumn{1}{c}{\textbf{Gemini}}
& \multicolumn{1}{c}{\cellcolor{tableshade}\textbf{\begin{tabular}[c]{@{}c@{}}FLUID-\\Agent\end{tabular}}}
& \multicolumn{1}{c}{\textbf{Human}} \\

\midrule

% ============================================================
% Reference Fidelity
% ============================================================

\multicolumn{13}{@{}l}{%
    \textit{Reference Fidelity}
} \\
Similarity F1 $\uparrow$
& \pmval{0.101}{0.018}
& \pmval{0.091}{0.016}
& \pmval{0.099}{0.014}
& \pmval{0.116}{0.021}
& \pmbest{0.146}{0.021}
& 1.000
& \pmval{0.045}{0.010}
& \pmval{0.048}{0.012}
& \pmbest{0.076}{0.012}
& \pmval{0.028}{0.008}
& \pmval{0.063}{0.012}
& 1.000 \\
LLM-Judged F1 $\uparrow$
& \pmval{0.127}{0.019}
& \pmval{0.138}{0.019}
& \pmval{0.138}{0.020}
& \pmval{0.127}{0.019}
& \pmbest{0.225}{0.022}
& 1.000
& \pmval{0.051}{0.010}
& \pmval{0.075}{0.015}
& \pmval{0.098}{0.015}
& \pmval{0.037}{0.013}
& \pmbest{0.099}{0.015}
& 1.000 \\
\midrule
% ============================================================
% Reactive Guidance
% ============================================================

\multicolumn{13}{@{}l}{%
    \textit{Reactive Guidance}
} \\
Timely Correctness $\uparrow$
& \pmval{0.088}{0.017}
& \pmval{0.095}{0.025}
& \pmval{0.107}{0.020}
& \pmval{0.126}{0.028}
& \pmbest{0.209}{0.030}
& \pmval{0.862}{0.027}
& \pmval{0.020}{0.013}
& \pmval{0.029}{0.013}
& \pmval{0.048}{0.017}
& \pmval{0.016}{0.012}
& \pmbest{0.055}{0.020}
& \pmval{0.615}{0.054} \\
Response Latency (s) $\downarrow$
& \pmval{2.00}{0.20}
& \pmval{1.70}{0.13}
& \pmval{1.52}{0.11}
& \pmval{2.16}{0.10}
& \pmbest{1.15}{0.04}
& \pmval{0.20}{0.05}
& \pmval{1.35}{0.40}
& \pmval{1.44}{0.19}
& \pmval{1.66}{0.38}
& \pmval{1.28}{1.43}
& \pmbest{1.17}{0.13}
& \pmval{0.47}{0.11} \\
\midrule
% ============================================================
% Proactive Guidance
% ============================================================

\multicolumn{13}{@{}l}{%
    \textit{Proactive Guidance}
} \\
Timely Correctness $\uparrow$
& \pmval{0.046}{0.022}
& \pmval{0.015}{0.012}
& \pmval{0.023}{0.012}
& \pmbest{0.063}{0.026}
& \pmval{0.046}{0.015}
& \pmval{0.921}{0.018}
& \pmval{0.025}{0.009}
& \pmval{0.018}{0.008}
& \pmval{0.029}{0.009}
& \pmval{0.020}{0.009}
& \pmbest{0.062}{0.013}
& \pmval{0.881}{0.011} \\
\midrule
% ============================================================
% Interaction Discipline
% ============================================================

\multicolumn{13}{@{}l}{%
    \textit{Intervention Precision and Turn-Taking}
} \\
Intervention Precision $\uparrow$
& \pmval{0.088}{0.016}
& \pmval{0.059}{0.012}
& \pmval{0.080}{0.014}
& \pmval{0.145}{0.023}
& \pmbest{0.164}{0.022}
& \pmval{0.845}{0.027}
& \pmval{0.052}{0.013}
& \pmval{0.073}{0.018}
& \pmval{0.050}{0.008}
& \pmval{0.128}{0.043}
& \pmbest{0.166}{0.024}
& \pmval{0.833}{0.021} \\
Speech Overlap Ratio $\downarrow$
& \pmval{0.213}{0.037}
& \pmval{0.442}{0.046}
& \pmval{0.224}{0.025}
& \pmbest{0.059}{0.008}
& \pmval{0.207}{0.014}
& \pmval{0.038}{0.009}
& \pmval{0.206}{0.032}
& \pmval{0.218}{0.039}
& \pmval{0.480}{0.071}
& \pmbest{0.063}{0.016}
& \pmval{0.153}{0.027}
& \pmval{0.116}{0.026} \\

\bottomrule

\end{tabular}%
}

\endgroup

\caption{Evidence-grounded evaluation on WTaG and HoloAssist. Values are reported with half-widths of 95\% session-cluster bootstrap confidence intervals.}
\label{tab:main-results}
\end{table*}

\subsection{Experimental Setup}
\label{sec:setup}

\paragraph{Systems.}
We evaluate four representative baselines with \Ours{}: ProAssist~8B \citep{zhang2025proactive}, VideoLLM-online~8B \citep{VideoLLM-online}, JoyAI-VL-Interaction \citep{joyai2026vlinteraction}, and Gemini 3.8 Live \citep{google2026gemini38live}. These baselines are selected based on four criteria:
(1) streaming input and output,
(2) proactive response generation,
(3) training on task-guidance data, and
(4) compatibility with the teacher-forced inference setting.

%\jycc{I would not position it as FlUID-agent against the others. Basically apply the evalution framework on a range of models/systems including FLUID-agent.} 
%\roihn{fixed}

\paragraph{Protocol.} 
% All systems receive the same structured task knowledge extracted from demonstration videos. We use teacher-forced inference, conditioning each system on the ground-truth interaction history rather than its own previous outputs. All systems operate in real time with streaming video and audio inputs and spoken audio output, allowing us to measure response latency and speech timing. See Appendix~\ref{app:setup} for implementation details.

To isolate the quality of individual responses in offline evaluation and avoid error propagation across the interaction, we use teacher-forced inference, conditioning each system on the ground-truth interaction history rather than its own previous outputs. All systems also receive the same structured task knowledge extracted from demonstration videos. They operate in real time with streaming video and audio inputs and spoken audio output, allowing us to measure response latency and speech timing. See Appendix~\ref{app:setup} for implementation details.
%\jycc{need to explain why teacher forcing? in order to  isolate individual responses in the offline evaluation. }
%\roihn{fixed}

\begin{figure*}[t]
    \centering
    \includegraphics[width=0.95\textwidth]{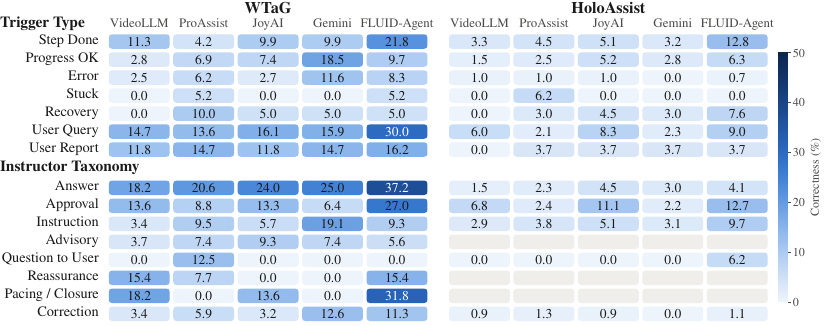}
    \caption{Fine-grained intervention correctness (\%).
Percentage of necessary interventions correctly addressed by each system, grouped by trigger type and instructor taxonomy. Gray cells indicate fewer than 10 interventions.}
    \label{fig:correctness-by-type}
\end{figure*}

%We evaluate whether models can provide context-aware guidance given task knowledge. To support this evaluation, we extract structured textual task knowledge from egocentric demonstration videos using a multi-round VLM pipeline. For WTaG, we derive shared task knowledge from one representative session per task; for HoloAssist, where execution procedures vary substantially across sessions, we construct session-specific task knowledge.

%With the structured task knowledge provided as context, we evaluate all baselines under a teacher-forced setting, conditioning each intervention on the ground-truth interaction history while excluding previous model outputs.

%\paragraph{Real-Time Execution.}
%We evaluate all systems in a real-time video-input, audio-output setting to reflect realistic perceptual guidance. Systems process streaming video causally and produce spoken responses, allowing us to measure end-to-end latency as well as actual speech onset and completion time, which cannot be faithfully captured from text outputs alone. 

\subsection{Main Results}
\label{sec:main-results}
We evaluate all models alongside the original human instructor guidance from both datasets under the FLUID-Evaluation framework. As shown in Table~\ref{tab:main-results}, human instructors substantially outperform all evaluated models, establishing a human performance baseline. By incorporating interaction dynamics, \Ours{} generally achieves more timely and precise interventions than other models. However, Gemini exhibits less speech overlap on both datasets, suggesting that effective turn-taking remains a challenge for \Ours{}. Substantial gaps remain between \Ours{} and human instructors, particularly in proactive Timely Correctness (below 0.07 vs. above 0.88) and Intervention Precision (below 0.17 vs. above 0.83). These results highlight the remaining challenges in providing timely and appropriate guidance.

\paragraph{Evidence-grounded evaluation captures intervention quality beyond reference matching.}
Systems with similar LLM-Judged F1 can exhibit substantially different interaction behavior. On WTaG, Gemini and VideoLLM achieve comparable F1 scores (0.127 vs.\ 0.138), yet Gemini has much higher Intervention Precision and substantially less speech overlap. Similarly, on HoloAssist, \Ours{} and JoyAI achieve nearly identical F1 (0.099 vs.\ 0.098), while \Ours{} achieves higher proactive Timely Correctness and Intervention Precision. By accounting for intervention necessity and evidence-derived timing windows, our evaluation reveals differences in intervention quality that reference-matching metrics cannot capture.

\begin{table*}[t]
\centering

\begingroup
\small
\setlength{\tabcolsep}{2.5pt}
\renewcommand{\arraystretch}{1.0}

% Highlight colors
\definecolor{onsetshade}{RGB}{232,241,250}
\definecolor{factshade}{RGB}{251,235,217}

\resizebox{\textwidth}{!}{%
\begin{tabular}{
    @{}l
    cccc
    ccc
    @{}
}

\toprule

\textbf{Metric}
& \textbf{L0} & \textbf{L1} & \textbf{L2} & \textbf{L3}
& $\Delta_{\mathrm{onset}}$
& $\Delta_{\mathrm{type}}$
& $\Delta_{\mathrm{fact}}$ \\

\midrule

\multicolumn{8}{@{}l}{\textbf{JoyAI}} \\

LLM-Judged F1 $\uparrow$
& \pmval{0.138}{0.020}
& \pmval{0.314}{0.034}
& \pmval{0.330}{0.035}
& \pmval{0.346}{0.038}
& \cellcolor{onsetshade}\pmval{+0.175}{0.029}
& \pmval{+0.016}{0.013}
& \pmval{+0.016}{0.013} \\

Reactive Timely Correctness $\uparrow$
& \pmval{0.107}{0.020}
& \pmval{0.375}{0.035}
& \pmval{0.409}{0.037}
& \pmval{0.411}{0.037}
& \cellcolor{onsetshade}\pmval{+0.269}{0.035}
& \pmval{+0.034}{0.020}
& \pmval{+0.002}{0.008} \\

Proactive Timely Correctness $\uparrow$
& \pmval{0.023}{0.012}
& \pmval{0.127}{0.033}
& \pmval{0.119}{0.034}
& \pmval{0.178}{0.035}
& \cellcolor{onsetshade}\pmval{+0.105}{0.032}
& \pmval{$-$0.008}{0.026}
& \cellcolor{factshade}\pmval{+0.059}{0.037} \\

Intervention Precision $\uparrow$
& \pmval{0.080}{0.014}
& \pmval{0.347}{0.037}
& \pmval{0.378}{0.040}
& \pmval{0.398}{0.037}
& \cellcolor{onsetshade}\pmval{+0.267}{0.035}
& \pmval{+0.031}{0.017}
& \pmval{+0.021}{0.015} \\

\midrule

\multicolumn{8}{@{}l}{\textbf{Gemini}} \\

LLM-Judged F1 $\uparrow$
& \pmval{0.127}{0.019}
& \pmval{0.301}{0.029}
& \pmval{0.313}{0.030}
& \pmval{0.362}{0.027}
& \cellcolor{onsetshade}\pmval{+0.174}{0.024}
& \pmval{+0.012}{0.023}
& \pmval{+0.049}{0.021} \\

Reactive Timely Correctness $\uparrow$
& \pmval{0.126}{0.028}
& \pmval{0.176}{0.027}
& \pmval{0.195}{0.028}
& \pmval{0.217}{0.031}
& \cellcolor{onsetshade}\pmval{+0.050}{0.030}
& \pmval{+0.019}{0.020}
& \pmval{+0.022}{0.026} \\

Proactive Timely Correctness $\uparrow$
& \pmval{0.063}{0.026}
& \pmval{0.259}{0.044}
& \pmval{0.273}{0.044}
& \pmval{0.354}{0.052}
& \cellcolor{onsetshade}\pmval{+0.195}{0.045}
& \pmval{+0.015}{0.036}
& \cellcolor{factshade}\pmval{+0.080}{0.043} \\

Intervention Precision $\uparrow$
& \pmval{0.145}{0.023}
& \pmval{0.291}{0.031}
& \pmval{0.310}{0.035}
& \pmval{0.348}{0.038}
& \cellcolor{onsetshade}\pmval{+0.145}{0.028}
& \pmval{+0.019}{0.026}
& \pmval{+0.038}{0.030} \\

\bottomrule

\end{tabular}%
}

\endgroup

\caption{
Evidence ladder on WTaG. L0 retains native behavior; L1 enforces interventions at evidence onsets; L2 adds trigger types; and L3 adds specific evidence.  $\Delta$ columns show differences (L1--L0, L2--L1, and L3--L2). Blue and orange shading highlight gains from onset control and specific evidence for proactive guidance, respectively.
}
\label{tab:evidence-ladder}
\end{table*}

\paragraph{Fine-grained evaluation reveals interaction-specific strengths and failures.}
Figure~\ref{fig:correctness-by-type} reports correctness by trigger type and instructor utterance type. \Ours{} leads across multiple categories on both datasets. Across systems, correctness is generally higher for explicit user queries, but remains low for proactive interventions requiring an understanding of task progress and user behavior. Correctness remains low for \textit{Error}  and \textit{Stuck} triggers, as well as \textit{Question to User} utterances that require seeking missing information. These results highlight the challenge of grounding proactive interventions in task evidence and deciding when and what to say, motivating our diagnostic analysis in Section~\ref{sec:evidence-bottlenecks}.

\subsection{Diagnostic Analysis with Evidence Ladder}
\label{sec:evidence-bottlenecks}

While the preceding results reveal difficulties in deciding when and what to say, it remains unclear how intervention decisions and evidence grounding contribute to these difficulties. To help better understand what types of evidence is difficult to capture and affects agent's performance, we construct an evidence ladder on WTaG that progressively controls intervention decisions and introduces ground-truth task evidence. L0 retains native behavior; L1 enforces one response at each annotated necessary-intervention onset while suppressing other speech; L2 adds the evidence type; and L3 further provides specific observed facts for action- and state-triggered events (see Appendix~\ref{app:ladder} for details). None of these levels provides the reference guidance. 
%\jycc{The goal of this analysis is to provide additional layer of ground-truth evidence from L0 up to L3 to help pin down the kind of evidence that is difficult to capture or affects agent's performance? Need to say why we are doing this. }\yunqi{fixed}

\paragraph{Deciding whether and when to intervene is a major bottleneck.}
The largest gains occur at L1 (Table~\ref{tab:evidence-ladder}), where reactive and proactive Timely Correctness increase by 0.269 and 0.105 for JoyAI, and by 0.050 and 0.195 for Gemini, respectively. Intervention Precision also improves, reflecting fewer unnecessary interventions. Together, these gains show the importance of deciding whether and when to intervene in real-time task guidance.

\paragraph{Specific evidence improves proactive guidance.}
Coarse evidence at L2 provides limited gains, as it lacks concrete details about the ongoing task. In contrast, providing specific observations at L3 improves proactive Timely Correctness by 0.059 for JoyAI and 0.080 for Gemini. Gains in reactive guidance are smaller, as user utterances already provide relevant evidence. These results suggest that stronger grounding in the ongoing task is key to effective proactive guidance.

\subsection{\Ours{} Ablation}
\label{sec:ablation}
We ablate the two decision modules in Figure~\ref{fig:framework} to assess taxonomy-guided reactive and evidence-grounded proactive decisions. \emph{Direct Reactive} replaces User Utterance Understanding with an LLM that directly decides whether and how to respond to each user utterance. \emph{Direct Proactive} replaces Intervention Decision with an LLM that directly decides whether to speak. Both retain the same backbone, perception pipeline, and task memory (see Appendix~\ref{app:ego-ablation} for details).

\paragraph{Taxonomy-guided understanding reduces unnecessary reactive responses.}
Compared with Direct Reactive, \Ours{} achieves comparable Reactive Timely Correctness on WTaG and higher correctness on HoloAssist, with higher Intervention Precision and fewer reactive utterances on both datasets. Direct Reactive responds more often to turns requiring no answer (see Response Rate on Unanswered Turns), supporting the role of taxonomy guidance in deciding when to respond and when to remain silent.

\begin{table}[t]
\centering
\small
\setlength{\tabcolsep}{3pt}
\renewcommand{\arraystretch}{1.02}
\resizebox{\linewidth}{!}{%
\begin{tabular}{@{}l >{\columncolor{tableshade}}c c >{\columncolor{tableshade}}c c@{}}
\toprule
& \multicolumn{2}{c}{\textbf{WTaG}}
& \multicolumn{2}{c}{\textbf{HoloAssist}} \\
\cmidrule(lr){2-3}\cmidrule(l){4-5}

\textit{(a) Utterance Understanding}
& \cellcolor{tableshade}\textbf{\begin{tabular}[c]{@{}c@{}}FLUID-\\Agent\end{tabular}}
& \textbf{\begin{tabular}[c]{@{}c@{}}Direct\\Reactive\end{tabular}}
& \cellcolor{tableshade}\textbf{\begin{tabular}[c]{@{}c@{}}FLUID-\\Agent\end{tabular}}
& \textbf{\begin{tabular}[c]{@{}c@{}}Direct\\Reactive\end{tabular}} \\
\midrule
Reactive Timely Correctness $\uparrow$
& 0.209 & 0.210 & 0.055 & 0.038 \\
Reactive Intervention Precision $\uparrow$
& 0.224 & 0.158 & 0.158 & 0.141 \\
Reactive Utterances/min
& 1.94 & 2.99 & 0.86 & 0.96 \\
Response Rate on Unanswered Turns $\downarrow$
& 0.236 & 0.426 & 0.316 & 0.405 \\
Speech Overlap Ratio $\downarrow$
& 0.207 & 0.294 & 0.153 & 0.168 \\

\midrule
\textit{(b) Intervention Decision}
& \cellcolor{tableshade}\textbf{\begin{tabular}[c]{@{}c@{}}FLUID-\\Agent\end{tabular}}
& \textbf{\begin{tabular}[c]{@{}c@{}}Direct\\Proactive\end{tabular}}
& \cellcolor{tableshade}\textbf{\begin{tabular}[c]{@{}c@{}}FLUID-\\Agent\end{tabular}}
& \textbf{\begin{tabular}[c]{@{}c@{}}Direct\\Proactive\end{tabular}} \\
\midrule
Proactive Timely Correctness $\uparrow$
& 0.046 & 0.119 & 0.062 & 0.171 \\
Proactive Intervention Precision $\uparrow$
& 0.072 & 0.020 & 0.171 & 0.049 \\
Proactive Utterances/min
& 1.32 & 14.83 & 1.55 & 16.88 \\
Repetition Rate $\downarrow$
& 0.095 & 0.619 & 0.089 & 0.684 \\
Speech Overlap Ratio $\downarrow$
& 0.207 & 0.752 & 0.153 & 0.799 \\
\bottomrule
\end{tabular}%
}
\caption{Ablation of \Ours{} decision modules. }
\label{tab:ablation}
\end{table}

\paragraph{Evidence-conditioned intervention decisions enable more selective proactive guidance.}
Compared with Direct Proactive, \Ours{} achieves higher Proactive Intervention Precision with roughly 11 times fewer proactive utterances and substantially less speech overlap on both datasets. Although Direct Proactive achieves higher Proactive Timely Correctness, over 60\% of its utterances repeat earlier responses (see Repetition Rate). These results show that structured intervention decisions enable more appropriate proactive guidance while reducing redundant and unnecessary speech.

\section{Discussion and Conclusion}
\label{sec:conclusion}
Recent advances in vision and language foundation models as well as agentic architecture have made it feasible to develop perceptual agents that can provide real-time task guidance to help people acquire new skills and complete physical tasks. However, evaluating perceptual guidance agents remains a significant challenge due to the inherent flexibility of instructional dialogue, where appropriate guidance can vary in content and timing. 

To address this issue, we developed FLUID, a framework that evaluates not only the accuracy of generated guidance but also the flexibility of AI agents in providing different types of guidance based on users’ needs and task contexts. FLUID grounds its evaluation metrics in human instructor guidance, focusing on the evidence supporting the guidance, its necessity, and the appropriateness of its timing.  Our current FLUID framework contains instructional interaction data, fine-grained annotation and evaluation protocols. It can be used to support systematic and comparative evaluation of AI agents’ guidance capabilities. Our experimental results show that, while human instructors achieve high performance under FLUID metrics, state-of-the-art AI models still lag significantly behind. These findings highlight the substantial gap between perceptual agents' capabilities and proficient human instruction. 

For tasks that require specialized procedures, such as special daily tasks relevant to different people, FLUID provides a pipeline to establish evaluation metrics based on a few examples of natural interaction sessions between an instructor and a follower.  FLUID can then provide an initial assessment of agent behavior and reliability before deploying agents in real-world user tasks.
%\roihn{To me this feels a bit unexpected. We do not mention how FLUID can guide people to collect data in our paper. A safer claim would be FLUID provides a pipeline that once some interaction data is collected, it can help to build up an evaluation framework for assessing and improving agents in specific fields.}

Evaluating real-time guidance systems is an extremely challenging problem. Ultimately, these agents must be evaluated through real-time interactions with actual users. However, such evaluations are prohibitively expensive and difficult to scale. FLUID provides a way to bootstrap the development and evaluation of agents’ guidance capabilities before moving to real-world user studies. We hope FLUID will offer new perspectives on how to evaluate real-time guidance systems and motivate the development of perceptual guidance agents capable of providing flexible, contextually appropriate, and timely guidance comparable to that of skilled human instructors.

%We propose FLUID, a framework for studying interaction flexibility from naturally occurring human-human task-guidance data. FLUID grounds interventions in their supporting evidence, necessity, and timing, and uses this representation to support both evaluation and real-time agent design. Our results show the value of moving beyond rigid reference matching toward more adaptive, evidence-grounded task guidance.

%\jycc{I think the goal of the framework is not about studying human-human task guidance, but rather developing a framework that allows better development and evaluation of AI agents that can provide flexible guidance as humans do. Imaging in the future, for any domains that require AI agents to provide situated task guidance, we can first collect a few examples of how humans insturctors teach human learners. Then we can use the collected data to evaluate AI agents performance and provide feedback to further develop and evaluate AI agents before their deployment to interact with real users.  }
\section*{Limitations}
\label{sec:limitations}

Our annotations primarily recover the grounded evidence underlying observed instructor interventions, which provides a strong reference for evaluation but does not exhaust all potentially valid interaction behaviors. In particular, the absence of an intervention in the demonstration does not imply that no intervention would have been appropriate at that moment. We also do not explicitly model silent moments, partly because WTaG and HoloAssist contain few cases where speaking is clearly prohibited; many additional utterances could be non-harmful as long as they do not disrupt task progress. Identifying such unobserved but potentially valid intervention opportunities, as well as moments where silence is truly required, involves more subjective judgments and is left for future work.

% \jycc{do we need to mention full-duplex?  If WTag adn HoloAssist based on human interaction data do not show full-duplex behaviors, I'm wondering maybe full-duplex, or whatever the definitions it holds, may not be that representative in the guidance tasks. I think we should remove the following paragraph. }
% Full-duplex interaction is another dimension of flexible interaction, but it is only sparsely represented in WTaG and HoloAssist and is not studied in this work. Our modular framework is built on a base model without native full-duplex capability, making overlapping perception and generation difficult to support naturally. Moreover, although such interactions exist, WTaG and HoloAssist contain relatively few cases involving frequent backchanneling, overlapping speech, or visual barge-ins. Future work could collect more interaction-intensive data with richer back-and-forth dynamics and extend the framework to full-duplex settings.

Our framework supports a broad range of proactive and reactive assistance behaviors in complex task interactions. However, more fluid and dynamic interaction scenarios, such as full-duplex perception and generation, remain challenging, as overlapping perception and response generation are difficult to support naturally within a modular system. In the datasets, we do observe several dynamic interactive behaviors, including backchanneling, overlapping speech, and visual barge-ins, and some of these are partially reflected in our evaluation. However, a comprehensive characterization and evaluation of richer full-duplex interaction is beyond the scope of this work. Future work could study these behaviors more systematically, supported by interaction data with denser and more diverse back-and-forth dynamics.
\section*{Acknowledgments}

This work was supported in part by the Johns Hopkins Artificial Intelligence and Technology Collaboratory (JH AITC), funded by the National Institute on Aging of the National Institutes of Health under grant P30AG073104. The content is solely the responsibility of the authors and does not necessarily represent the official views of the National Institutes of Health.

% \clearpage

% \bibliography{\NAACLdir references}
\bibliography{references}

\clearpage
\appendix
\section*{Appendix}

% Appendix contents: the sections and subsections of the appendix with their page numbers.
% The invisible part-level entry makes the local contents list below cover everything that
% follows it, so the main text and the references are excluded. It also groups the appendix
% sections under an "Appendix" PDF bookmark. Use \etocsetnexttocdepth{section} to list
% sections only.
\phantomsection
\addcontentsline{toc}{part}{Appendix}
\begingroup
  \small
  \etocsettocstyle{\subsection*{Contents}}{}
  \etocsetnexttocdepth{subsection}
  \localtableofcontents
\endgroup

% \let\NAACLsection\section             \renewcommand{\section}{\par\allowbreak\NAACLsection}
% \let\NAACLsubsection\subsection       \renewcommand{\subsection}{\par\allowbreak\NAACLsubsection}
% \let\NAACLsubsubsection\subsubsection \renewcommand{\subsubsection}{\par\allowbreak\NAACLsubsubsection}

% \newcommand{\appendixfile}[1]{\IfFileExists{appendix/#1.tex}{\input{appendix/#1}}{}}
% \appendixfile{A_dataset_preparation}
% \appendixfile{B_annotation_protocol}
% \appendixfile{C_evaluation_protocol}
% \appendixfile{D_egoguide_framework}
% \appendixfile{E_experimental_setup}
% \appendixfile{F_evidence_ladder}
% \appendixfile{G_human_timing_study}

\section{Dataset Construction and Preprocessing}
\label{app:dataset}

WTaG and HoloAssist record task guidance in which a remote instructor
observes the user's view and provides verbal assistance. This appendix
describes how sessions were selected from the two datasets, how their
video, audio, and transcripts were prepared, and how the resulting
sessions differ in their interaction.

\subsection{Source Datasets and Session Selection}

From WTaG, we exclude sessions with incomplete task executions. From
HoloAssist, we select sessions with substantive user--instructor
interaction, excluding those dominated by repetitive next-step
instructions and acknowledgments, and then manually check the selected
recordings for audio quality. Table~\ref{tab:data-stats} reports the
final dataset sizes.

% Keep the shared statistics table here, after the selection description.
% This local wrapper preserves its contents, caption, and label without floating.
\begingroup
\renewenvironment{table}[1][]{%
  \par\addvspace{\intextsep}\noindent
  \begin{minipage}{\columnwidth}
  \captionsetup{type=table,hypcap=false}%
}{%
  \end{minipage}\par\addvspace{\intextsep}%
}
\begin{table}[t]
\centering
\small
\begin{tabular}{@{}lrr@{}}
\toprule
& \textbf{WTaG} & \textbf{HoloAssist} \\
\midrule
Video hours & 7.57 & 6.29 \\
Test sessions & 43 & 72 \\
User utterances & 2,324 & 1,358 \\
Instructor utterances & 1,482 & 2,736 \\
\bottomrule
\end{tabular}
\caption{Statistics of the two source datasets. HoloAssist instructor
utterances are counted as native transcript rows; merging fragments of
one contribution gives 2,731 units.}
\label{tab:data-stats}
\end{table}

\endgroup

\subsection{Multimodal Preprocessing and Synchronization}
\label{app:multimodal-preprocessing}

Video, audio, and annotation timestamps are measured in seconds from the
start of each recording. Video is sampled at 8\,Hz using presentation
timestamps, taking the latest frame available at each sampling time.

For the transcripts, we manually revise the WTaG transcripts to correct
speech boundaries and restore missing utterances, and keep HoloAssist's
released transcripts and timing. These speaker-labeled transcripts are
used for annotation and evaluation; at runtime, user text comes from
streaming automatic speech recognition (ASR) instead.

For the audio, instructor speech is muted, except where it overlaps user
speech. There it is left in place to preserve the user's signal, so the
instructor may remain audible in overlap regions. Recordings are replayed
continuously. Streaming ASR produces partial transcripts during an
utterance and a final transcript once the utterance ends. These outputs
become available only after the corresponding audio has been received and
processed.

\subsection{Dataset Statistics and Interaction Characteristics}
\label{app:dataset-statistics}

HoloAssist sessions are shorter than WTaG sessions but contain denser
instructor guidance. The median session duration is 265.8\,s in HoloAssist
and 594.0\,s in WTaG, while instructors speak 8.25 and 3.21 turns per
minute on average per session, with a median interval between adjacent
instructor utterances of 1.99\,s and 7.19\,s, respectively.

\begin{figure*}[t]
  \centering
  \includegraphics[width=\textwidth]{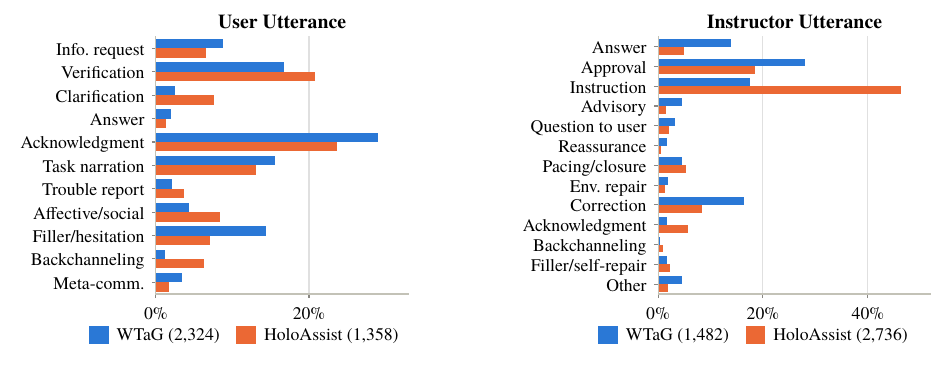}
  \caption{Utterance taxonomy distributions in WTaG and HoloAssist.}
  \label{fig:annotation-taxonomy}
\end{figure*}

The two datasets also differ in the type of guidance
(Figure~\ref{fig:annotation-taxonomy}). HoloAssist has a higher
proportion of instructions, while WTaG has higher proportions of answers,
approvals, and corrections. This pattern is consistent with denser
ongoing instruction in HoloAssist and more frequent answers and
corrections in WTaG. Correction labels describe the instructor's guidance, not the
frequency of actual task errors.

The necessary interventions in the released annotations show a
corresponding difference in how guidance is initiated. In WTaG, 66.1\% are
reactive and 33.9\% proactive; in HoloAssist, 31.4\% and 68.6\%. WTaG
therefore puts more weight on responding to users, while HoloAssist more
often requires guidance that is not prompted by a user request.

% Requires graphicx, booktabs, array, and tabularx.
\section{Evidence-Grounded Annotation Protocol}
\label{app:annotation}

The annotations connect each instructor intervention to the user speech,
visible actions, and task states that support it. This appendix defines
the four annotation layers, including the rules for intervention
necessity and timing, then describes the annotation workflow and the
review procedures for WTaG and HoloAssist, and closes with boundary cases
and a worked example.

\subsection{Annotation Overview and Label Schema}

We annotate 43 WTaG and 72 HoloAssist test sessions with the four linked
layers of Section~\ref{sec:interaction-structure}.
\emph{Layer 1: Utterance Taxonomy} assigns each turn a communicative-function
label and at most one attribute. \emph{Layer 2: Action and State} locates
visible task evidence in time and records its validity and task effect.
\emph{Layer 3: Trigger Link} connects each instructor intervention to its
supporting evidence and records its trigger type and intervention status.
\emph{Layer 4: Evidence Onset and Intervention Window} records the temporal
anchors that define when the intervention should begin.
Three additional WTaG sessions provide task knowledge only
and are not annotated or evaluated.

The unit of annotation for speech is the \emph{utterance unit}, one
speaker contribution in the cleaned transcript. Fragments of one
contribution that were split by speech recognition or transcript cleaning
are merged before labeling. A contribution is split into several units
only when its functions are clearly separable, for example \textit{There
is already water in it.} (U7) and \textit{Should I pour it out?} (U2);
otherwise it receives the label of its main communicative function.

\subsection{Layer 1: Utterance Taxonomy}

This layer labels the communicative function of each utterance, that is,
what the speaker is asking, reporting, or responding to. Users and
instructors have separate label sets, and labels are assigned from the
transcript and the conversational context. For brief responses, the
position of the turn in the interaction also matters, since it separates
acknowledgments from backchannels (U5 vs.\ U10; I10 vs.\ I11).

\subsubsection{User Utterance Taxonomy}

Eleven labels describe user turns: requests, responses, task reports, and
speech with little or no task-related content
(Table~\ref{tab:user_taxonomy}). When more than one label could apply,
annotators check the labeling rules in the order U10, U9, U11, U8, U3, U2,
U1, U4, U5, U7, U6 and use the first one that fits. An incomplete
utterance that clearly asks a question is not labeled U9 but receives one
of the three question labels: U3 asks about something the instructor
recently said, U2 proposes an answer about the task or the world and asks
whether it is correct, and U1 asks for information without proposing an
answer. Appendix~\ref{app:utterance_decisions} explains how these labels
are separated when the interpretation depends on context.

\begin{table*}[t]
\centering
\small
\setlength{\tabcolsep}{5pt}
\renewcommand{\arraystretch}{1.12}
\begin{tabularx}{\textwidth}{@{}>{\raggedright\arraybackslash}p{0.05\textwidth}>{\raggedright\arraybackslash}X@{}}
\toprule
\textbf{ID} & \textbf{Label, definition, and example} \\
\midrule
U1 & \textbf{Information request}. An open request for task content the user lacks; no candidate answer is
proposed. \textit{Umm, how much oil?} \\
U2 & \textbf{Verification question}. Proposes a candidate object, amount, action, plan, interpretation, state,
or result and requests validation, as a question or as a declarative
awaiting confirmation. \textit{It is a teaspoon, right?} \\
U3 & \textbf{Clarification request}. Requests repetition, rephrasing, or confirmation of the user's hearing or
understanding of an instructor utterance. \textit{What do you mean by
pinch, sorry?} \\
U4 & \textbf{Answer}. Supplies information requested by an instructor question, state check, or
check instruction. \textit{Five.} \\
U5 & \textbf{Acknowledgment/\allowbreak uptake}. Signals receipt, understanding, or acceptance of a completed instructor
turn without new task content, including an echo the user closes
themselves. \textit{Twelve ounces? Okay.} \\
U6 & \textbf{Task narration}. Propositional verbalization of the user's own action, plan, reasoning,
need, decision, or progress, including reading instructions aloud.
\textit{I already added the sugar.} \\
U7 & \textbf{Trouble report}. Reports that something is wrong or has deviated from the expected course.
\textit{Wait, that wasn't a tablespoon.} \\
U8 & \textbf{Affective expression/\allowbreak social talk}. Emotion, evaluation, politeness, apology, laughter, or conversation not
required for task completion. \textit{Oh my God.} \\
U9 & \textbf{Filler/\allowbreak hesitation}. Non-propositional or self-directed speech: fillers, false starts,
fragments, muttered reading, self-directed pacing tokens, or
unintelligible speech. \textit{And, uh\ldots} \\
U10 & \textbf{Backchanneling}. A brief continuer produced while the instructor is still speaking.
\textit{Mm-hm.} \\
U11 & \textbf{Meta-communication}. Discusses or manages the interaction or session itself, such as protocol,
recording, apparatus, or audio and camera tests. \textit{Can you still hear
me?} \\
\bottomrule
\end{tabularx}
\caption{User utterance taxonomy. Examples illustrate a label but do not
replace the context-dependent rules in
Appendix~\ref{app:utterance_decisions}.}
\label{tab:user_taxonomy}
\end{table*}

\subsubsection{Instructor Utterance Taxonomy}

Thirteen labels describe instructor turns
(Table~\ref{tab:instructor_taxonomy}). They record what the instructor is
doing in the conversation, regardless of whether the content is correct.
When more than one label could apply, annotators check the labeling rules
in the order I12, I13, I8, I5, I9, I6, I2, I1, I4, I3, I7, I11, I10 and
use the first one that fits. A correction is therefore labeled as a
correction even if it also expresses approval or answers a question, and
information given in response to a user request is labeled as an answer
rather than as an instruction offered without a request. For turns with
little substantive content, the label depends on timing: I11 while the
user is still speaking or performing an action, and I10 after the user
has finished.

\begin{table*}[t]
\centering
\small
\setlength{\tabcolsep}{5pt}
\renewcommand{\arraystretch}{1.12}
\begin{tabularx}{\textwidth}{@{}>{\raggedright\arraybackslash}p{0.05\textwidth}>{\raggedright\arraybackslash}X@{}}
\toprule
\textbf{ID} & \textbf{Label, definition, and example} \\
\midrule
I1 & \textbf{Answer}. Supplies task content the user requested explicitly or implicitly, through
a question, clarification request, stated need, trouble report, or audible
search. \textit{The filter cone is the metal thing.} \\
I2 & \textbf{Approval}. Validates a concrete user candidate, action, state, or result as correct or
acceptable. \textit{Yep, like that.} \\
I3 & \textbf{Instruction}. Introduces or elaborates a required task step on the instructor's own
initiative, without presupposing an error. \textit{Grind it for about twenty
seconds.} \\
I4 & \textbf{Advisory}. Gives optional, cautionary, conditional, or alternative guidance that does
not repair an existing error. \textit{You can use the glass bowl if that is
easier.} \\
I5 & \textbf{Question-to-user}. Requests information about the task state, the user's intent, the user's
speech, or the user's understanding. \textit{How many toothpicks do you
have?} \\
I6 & \textbf{Reassurance}. Mitigates user worry, trouble, or a mishap by marking it as tolerable or
recoverable. \textit{That is okay, no worries.} \\
I7 & \textbf{Pacing/\allowbreak closure}. Regulates starting, continuing, waiting, transitioning, or ending without
substantive procedural content. \textit{Keep going.} \\
I8 & \textbf{Environmental repair}. Asks the user to improve the instructor's incoming visual or audio
evidence. \textit{Could you look down more?} \\
I9 & \textbf{Correction}. Repairs an actual, proposed, imminent, or completed user error or an
omitted step. \textit{No, use the jelly instead of the marmalade.} \\
I10 & \textbf{Acknowledgment}. Signals receipt of a completed user contribution without information,
evaluation, reassurance, or flow direction. \textit{Oh, okay.} \\
I11 & \textbf{Backchanneling}. A brief, low-content signal of attention while the user is still speaking
or performing an extended action. \textit{Mm-hm.} \\
I12 & \textbf{Filler/\allowbreak self-repair}. Hesitation, an abandoned start, or a repair of the instructor's own wording
or instruction. \textit{Sorry, I meant one teaspoon.} \\
I13 & \textbf{Other}. Meta-communication, session or recording management, connection
checks, social talk, or rare residual content. \textit{You can stop the
recording.} \\
\bottomrule
\end{tabularx}
\caption{Instructor utterance taxonomy.}
\label{tab:instructor_taxonomy}
\end{table*}

\subsubsection{Utterance Attributes}

An attribute records what a turn is about, while the function label
records what the turn does. Each turn takes at most one category-specific
attribute (Table~\ref{tab:utterance_attributes}); U4, U5, U9--U11, I6, I10,
and I13 take none. U2, I1--I4, and I9 share
\texttt{content\_\allowbreak target}, whose values are
\texttt{object\_\allowbreak tool}, \texttt{action\_\allowbreak method},
\texttt{quantity\_\allowbreak parameter}, \texttt{sequence\_\allowbreak step},
\texttt{state\_\allowbreak result}, and \texttt{other\_\allowbreak unclear},
so that a verification and its response can be compared on the same
content axis. Figure~\ref{fig:overview} shows the label and attribute
together, such as Verification--quantity.

\begin{table}[tb]
\centering
\small
\setlength{\tabcolsep}{3pt}
\renewcommand{\arraystretch}{1.08}
\begin{tabularx}{\columnwidth}{@{}l>{\raggedright\arraybackslash}X@{}}
\toprule
\textbf{Labels} & \textbf{Attribute and focus} \\
\midrule
U1 & \texttt{request\_\allowbreak type}: information requested. \\
U2, I1--I4, I9 & \texttt{content\_\allowbreak target}: task content. \\
U3 & \texttt{repair\_\allowbreak target}: hearing, meaning, or task detail. \\
U6 & \texttt{narration\_\allowbreak type}: action, progress, reasoning, or reading aloud. \\
U7 & \texttt{trouble\_\allowbreak type}: problem reported. \\
U8 & \texttt{expression\_\allowbreak type}: affective, task-related, or social talk. \\
I5 & \texttt{question\_\allowbreak purpose}: information sought. \\
I7 & \texttt{flow\_\allowbreak function}: starting, continuing, waiting, transition, or closure. \\
I8 & \texttt{repair\_\allowbreak target}: visual or audio access. \\
I11 & \texttt{backchannel\_\allowbreak target}: speech, action, progress, or affect/trouble. \\
I12 & \texttt{repair\_\allowbreak form}: filler or self-repair. \\
\bottomrule
\end{tabularx}
\caption{Utterance attributes. The two uses of
\texttt{repair\_\allowbreak target} have different value sets.}
\label{tab:utterance_attributes}
\end{table}

\subsection{Layer 2: Action and State}
\label{app:action_schema}

Annotators identify actions and task states from visible evidence in the
video, using the dialogue as additional context when needed, and mark the
start and end time of each. Each interval is written as
$[t_{\mathrm{start}},t_{\mathrm{end}})$, which includes the start time and
excludes the end time. The interval of an action is called an action
span, and each span also receives a validity label and a task-effect
label, defined below. Table~\ref{tab:action_schema} lists the fields
recorded for each WTaG annotation.

\begin{table}[tb]
\centering
\small
\setlength{\tabcolsep}{3pt}
\renewcommand{\arraystretch}{1.08}
\begin{tabularx}{\columnwidth}{@{}>{\raggedright\arraybackslash}p{0.23\columnwidth}>{\raggedright\arraybackslash}X@{}}
\toprule
\textbf{Field} & \textbf{Values or boundary rule} \\
\midrule
Unit kind &
\texttt{action} (motion directed toward a task goal) or
\texttt{state\_\allowbreak event} (a task condition or a change
in that condition). \\
Times &
Start: the first frame that clearly shows the action or state.
End: the first frame after it ends or changes identity. \\
Verb/noun &
A short description supported by the video,
e.g., \texttt{grab--fork}. \\
Validity &
\texttt{VALID}, \texttt{DEVIATION}, \texttt{AMBIGUOUS}. \\
Task effect &
\texttt{NO\_\allowbreak PROGRESS}, \texttt{PROGRESS},
\texttt{COMPLETION}, \texttt{DEVIATION\_\allowbreak ONSET},
\texttt{RECOVERY}, \texttt{UNKNOWN}. \\
State label &
For \texttt{state\_\allowbreak event} units: \texttt{STALL},
\texttt{SEARCHING}, \texttt{VISIBILITY\_\allowbreak LOSS},
\texttt{VISIBILITY\_\allowbreak RECOVERY},
\texttt{SALIENT\_\allowbreak STATE\_\allowbreak CHANGE}. \\
\bottomrule
\end{tabularx}
\caption{Fields used to annotate actions and task states in WTaG.}
\label{tab:action_schema}
\end{table}

An action span covers one continuous attempt to achieve a task goal. It
may include several hand movements as long as the goal, object, and
validity stay the same; a new span starts when any of these changes, or
when the user begins another attempt after disengaging their hand or
resetting the task state. If an action is briefly occluded, the span
continues when continuous motion lets us identify it across the
occlusion. If the action cannot be identified reliably, the uncertain
interval is left unannotated.

Validity indicates whether the observed behavior is appropriate given the
task instructions and the current state: \texttt{VALID} when it follows
these requirements, \texttt{DEVIATION} when it conflicts with a task goal
or constraint, and \texttt{AMBIGUOUS} when the available evidence is
insufficient to decide.

Task effect records how the action changes the task state.
\texttt{PROGRESS} marks an action that advances a step or subgoal, and
\texttt{COMPLETION} one that completes a step or subgoal or reaches a
meaningful task state. \texttt{DEVIATION\_ONSET} identifies the action
that first creates an erroneous or undesirable state, and
\texttt{RECOVERY} an action that begins or completes the correction of
that state. \texttt{NO\_PROGRESS} marks an action with no substantive
advancement, and \texttt{UNKNOWN} one whose effect cannot be determined
reliably. A span can carry several task-effect labels if the task state
changes during the same continuous action, for example when an action
first makes progress and then completes a step.

\subsubsection{Action Vocabulary}
\label{app:action_vocab}

WTaG action descriptions use task-specific vocabularies developed from
the recipes and checked against the videos. Qwen3-VL-30B-A3B-Instruct
generated initial verb, noun, and adjective lists from the task steps,
their execution variants, and the 48 HoloAssist verbs as a reference, and
then checked sampled frames from 506 video segments for missing words.
Annotators reviewed the words supported by at least two segments and
merged synonyms. Compound objects are represented by a noun and an
adjective, such as \texttt{jar} with \texttt{nut\_\allowbreak butter};
locations and hand use are recorded separately. A final video check with
Qwen3-VL-32B-Instruct added \texttt{remove} and \texttt{shake} and found
no missing nouns. The resulting vocabulary contains 83 verbs, 120 nouns,
and 107 adjectives. Observations outside the vocabulary are retained as
out-of-vocabulary descriptions rather than forced into an unsuitable
category.

\subsubsection{Action-Candidate Generation and Human Review}
\label{app:action_candidates}

WTaG action annotations combine model proposals with a human scan of the
full task timeline. Four models independently propose intervals and
descriptions: Qwen3.5-27B, Qwen3.6-35B-A3B, Qwen3.8-27B, and
Qwen3-VL-32B-Instruct. Temporally clustered proposals supported by at
least two models enter review, including incompatible proposals that
meet this threshold; the votes decide only which candidates are reviewed,
not their validity or boundaries. Reviewers resolve the candidates using
the video and task context and add manual \emph{pins} for missed events,
which follow the same review procedure. Any candidate may be accepted,
edited, merged, split, or rejected: of 2,611 model proposals, 665 were
kept unchanged, 1,423 were kept after editing, and 523 were rejected.
After this review, a separate model-based pass assigns validity
and task effect to the reviewed actions using the task context and the
annotation guidelines.

\subsection{Layer 3: Trigger Link}
\label{app:trigger_schema}

For every instructor utterance, the annotator selects the smallest set
of preceding evidence that warrants the intervention. A link records the instructor
utterance ID, the evidence source, one or more evidence IDs, the trigger
type, a confidence, and optional ambiguity flags;
Table~\ref{tab:trigger_schema} gives the controlled values.

\begin{table}[tb]
\centering
\small
\setlength{\tabcolsep}{3pt}
\renewcommand{\arraystretch}{1.08}
\begin{tabularx}{\columnwidth}{@{}>{\raggedright\arraybackslash}p{0.23\columnwidth}>{\raggedright\arraybackslash}X@{}}
\toprule
\textbf{Field} & \textbf{Controlled values} \\
\midrule
Source & \texttt{USER\_\allowbreak UTTERANCE},
\texttt{USER\_\allowbreak ACTION\_\allowbreak STATE}, \texttt{MIXED},
\texttt{INSTRUCTOR\_\allowbreak SELF}, \texttt{NONE\_\allowbreak FOUND}. \\
Trigger type & \texttt{USER\_\allowbreak QUERY}, \texttt{USER\_\allowbreak REPORT},
\texttt{ERROR}, \texttt{PROGRESS\_\allowbreak OK}, \texttt{STEP\_\allowbreak DONE},
\texttt{STUCK}, \texttt{RECOVERY}, \texttt{VISIBILITY},
\texttt{UNCLEAR\_\allowbreak STATE}, \texttt{DIALOG\_\allowbreak OTHER},
\texttt{SELF}, \texttt{NONE}. \\
Confidence & \texttt{low}, \texttt{medium}, \texttt{high}. \\
Audit flags & \texttt{unclear\_\allowbreak source},
\texttt{competing\_\allowbreak evidence}, \texttt{missing\_\allowbreak evidence},
\texttt{boundary\_\allowbreak uncertain}. \\
\bottomrule
\end{tabularx}
\caption{Trigger-link schema. Confidence concerns the evidence identity
and relation, not transcription quality.}
\label{tab:trigger_schema}
\end{table}

The source is \texttt{MIXED} only when neither modality alone is
sufficient. For example, \textit{Does that look good?} asked while the
user is rolling the tortilla links the instructor's \textit{Yeah} to both the U2 turn and the
visible action. An intervention is \emph{reactive} if its trigger
evidence includes a completed user utterance
(\texttt{USER\_\allowbreak UTTERANCE} or \texttt{MIXED}) and
\emph{proactive} otherwise, so this intervention is reactive. Whether an
intervention is reactive is determined by its trigger link, not its
utterance label: an I1
Answer is proactive when it responds to a visible search without a user
turn, such as searching for a half-teaspoon.

\subsubsection{Intervention Necessity}
\label{app:necessity}

An intervention is \emph{necessary} when the user needs it to avoid a
mistake, get unstuck, or obtain needed information; \emph{optional} when
it can be omitted while the user proceeds correctly; and \emph{excluded}
when it is task-irrelevant or requires knowledge outside the shared view,
the dialogue, or the task knowledge. Table~\ref{tab:necessity_rules}
assigns these statuses from the trigger type, the linked user label, and
the instructor label. When a link has several trigger types, the highest
status applies, in the order necessary $>$ optional $>$ excluded. Only
necessary interventions are scored by Correctness and Timely Correctness;
optional interventions take part in matching
(Section~\ref{sec:eval-matching}) but are not scored.

\begin{table*}[t]
\centering
\small
\setlength{\tabcolsep}{5pt}
\renewcommand{\arraystretch}{1.12}
\begin{tabularx}{\textwidth}{@{}>{\raggedright\arraybackslash}p{0.34\textwidth}>{\raggedright\arraybackslash}X>{\raggedright\arraybackslash}p{0.12\textwidth}@{}}
\toprule
\textbf{Trigger type} & \textbf{Condition} & \textbf{Status} \\
\midrule
Any &
Relies on knowledge that cannot be obtained from the shared egocentric view, the dialogue, or the task knowledge & Excluded \\
\texttt{USER\_\allowbreak QUERY}, \texttt{ERROR},
\texttt{UNCLEAR\_\allowbreak STATE}, \texttt{VISIBILITY}, \texttt{STUCK} &
None & Necessary \\
Any &
Linked user utterance is U1, U2, U3, or U7 & Necessary \\
\texttt{STEP\_\allowbreak DONE}, \texttt{PROGRESS\_\allowbreak OK},
\texttt{RECOVERY}, \texttt{USER\_\allowbreak REPORT} &
Instructor label is I1, I3, I8, or I9 & Necessary \\
\texttt{STEP\_\allowbreak DONE}, \texttt{PROGRESS\_\allowbreak OK},
\texttt{RECOVERY}, \texttt{USER\_\allowbreak REPORT} &
Any other instructor label & Optional \\
\texttt{DIALOG\_\allowbreak OTHER} & None & Optional \\
\texttt{NONE}, \texttt{SELF} & None & Excluded \\
\bottomrule
\end{tabularx}
\caption{Rules for intervention status.}
\label{tab:necessity_rules}
\end{table*}

For example, \textit{Yep} in response to \textit{Twelve ounce right?}
(U2) is necessary, while \textit{Great} after a correct pour
(\texttt{PROGRESS\_\allowbreak OK}, I2) is optional
(Figure~\ref{fig:overview}). On WTaG, the rules assign 1,158 of 1,482
statuses (78.1\%), 121 model proposals are retained after review
(8.2\%), and annotators set 203 (13.7\%); every status is reviewed. On
HoloAssist, Qwen3.8-27B assigns 2,492 of 2,736 statuses (91.1\%), the
rules assign 235 (8.6\%), and annotators set 9 (0.3\%).

\subsection{Layer 4: Evidence Onset and Intervention Window}
\label{app:timing_definitions}

Temporal anchors define when the \emph{target instructor intervention}
can begin. They are annotated against a streaming prefix: evidence after
time $t$ may confirm that a condition continues and may identify a later
deadline, but may not be used to claim that enough evidence was already
available at $t$. In the main text, ES, EO, LU, and R are written
$t_{\mathrm{early}}$, $t_{\mathrm{evid}}$, $t_{\mathrm{latest}}$, and
$t_{\mathrm{usr}}$.

\paragraph{Action-Triggered Interventions.}
Proactive interventions use three anchors. \textbf{Evidence Start (ES)} is
the earliest frame showing a specific sign that the intervention may be
needed, even if the condition is not yet certain; entering a task step
alone does not establish ES. \textbf{Evidence Onset (EO)} is the earliest
frame that makes the condition unambiguous and complete enough to justify
the intervention; the whole object need not be visible, but its identity
and task relation must be supported. \textbf{Latest Useful Time (LU)} is
the exclusive boundary after which the target intervention is no longer
useful on the recorded trajectory. The valid window is
$[\mathrm{ES},\mathrm{LU})$, with the timing score peaking at EO
(Section~\ref{sec:eval-timing}).

\paragraph{Utterance-Triggered Interventions.}
For a reactive intervention, including one with \texttt{MIXED} evidence,
the first eligible time is the \textbf{Response-Ready Time (R)}, the end
of the completed user turn in the trigger evidence; ES and EO are not
separately annotated for utterance-triggered interventions. Partial speech does not set R unless the utterance
is already semantically complete and only a delayed transcription
endpoint remains. The reactive window is $[\mathrm{R},\mathrm{LU})$.

\paragraph{Operational Definition of LU.}
LU is the earlier of task expiry and dialogue cutoff on the recorded
trajectory:
\[
  \mathrm{LU} =
  \min(\mathrm{task\mbox{-}expiry},\ \mathrm{dialogue\mbox{-}cutoff}).
\]
Task expiry occurs when the user completes the relevant action, starts
self-correction or recovery, or reaches a state in which the intervention
would have to be reformulated. Dialogue cutoff is the start of the first
acknowledgment, answer, or downstream action that presupposes the target
intervention, since inserting the intervention later would reverse the
recorded conversational order. Unrelated speech does not set LU. For
self-correction, LU marks its start rather than its completion.

\paragraph{Frame Selection and Precision.}
Annotators scrub coarsely, then step frame by frame around each boundary
and select the earliest frame that satisfies the definition. Stored
timestamps are in seconds at the source video's frame resolution and are
not rounded to whole seconds.

\subsection{Annotation Workflow and Quality Control}
\label{app:annotation_workflow}

Annotation proceeds layer by layer. Annotators first review
segmentation, transcripts, and utterance labels, then action candidates
and manual pins, and finally trigger links, intervention statuses, and
temporal anchors. Completed upstream layers stay visible during review,
and errors found in them are resolved with the annotator responsible.
The three annotators are authors and received no additional
compensation. They developed the rubrics together, annotated a shared
session, and discussed disagreements before annotation began, and
guideline changes led to re-review of the affected annotations. Review
coverage differs between the two datasets (Table~\ref{tab:provenance}).

\begin{table*}[t]
\centering
\small
\setlength{\tabcolsep}{5pt}
\renewcommand{\arraystretch}{1.12}
\begin{tabularx}{\textwidth}{@{}>{\raggedright\arraybackslash}p{0.28\textwidth}>{\raggedright\arraybackslash}X>{\raggedright\arraybackslash}X@{}}
\toprule
\textbf{Layer} & \textbf{WTaG (43 sessions)} & \textbf{HoloAssist (72 sessions)} \\
\midrule
1. Utterance Taxonomy &
Qwen3.6-35B-A3B; all reviewed &
Consensus of four models\textsuperscript{a}; audited \\
2. Action and State &
Four models\textsuperscript{a}, $\geq$2 votes, plus manual pins; all
reviewed &
Original dataset actions and model-proposed state events; not reviewed \\
3. Trigger Link &
Links: Qwen3.8-27B. Statuses: rules
(Table~\ref{tab:necessity_rules}) and annotators. All reviewed. &
Links: merged proposals of four models\textsuperscript{a}.
Statuses: Qwen3.8-27B and rules. Audited. \\
4. Evidence Onset and Intervention Window &
Model-proposed; all reviewed &
Model-proposed, as for WTaG; audited \\
\bottomrule
\end{tabularx}
\caption{Annotation provenance for the four layers.
\textsuperscript{a}Qwen3.5-27B,
Qwen3.6-35B-A3B, Qwen3.8-27B, and Qwen3-VL-32B-Instruct. ``Audited'':
reviewed in eight sessions after release
(Appendix~\ref{app:timing_qc}).}
\label{tab:provenance}
\end{table*}

Before release, schema, ordering, and reference checks are run and the
flagged cases are resolved. The automatic checks reject unknown
controlled values, missing referenced evidence, non-positive intervals,
and impossible timing orders: proactive cases require
$\mathrm{ES}\leq\mathrm{EO}$ and $\mathrm{ES}<\mathrm{LU}$, and reactive
cases require $\mathrm{R}<\mathrm{LU}$. Human-audit flags cover competing
triggers, missing evidence, and uncertain boundaries.

\subsubsection{Review Coverage and Audit}
\label{app:timing_qc}

\begin{figure}[t]
\centering
\includegraphics[width=\columnwidth]{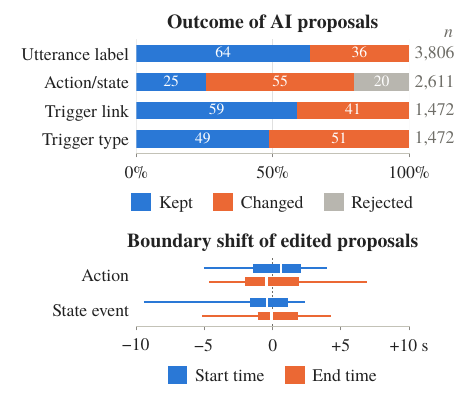}
\caption{\textbf{WTaG manual review.} Top: review outcome of the AI
proposals in the 43 test sessions; $n$ is the number of proposals per
layer. Bottom: shift of the edited start and end times of the retained
action and state proposals (10th, 25th, 50th, 75th, and 90th
percentiles).}
\label{fig:annotation-wtag-edits}
\end{figure}

\begin{figure}[t]
\centering
\includegraphics[width=\columnwidth]{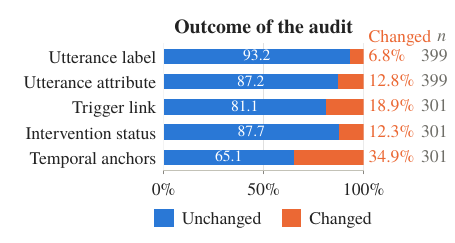}
\caption{\textbf{HoloAssist audit.} Annotation changes in the
eight-session audit. $n$ is 399 utterances for labels and attributes and
301 trigger links for the other fields.}
\label{fig:annotation-holoassist-errors}
\end{figure}

All layers of all 43 WTaG test sessions were manually reviewed.
Figure~\ref{fig:annotation-wtag-edits} summarizes the changes to the WTaG
drafts and to action/state boundaries: annotators adjusted 68.1\% of the
proposed ES anchors, 78.6\% of the EO anchors, and 56.1\% of the LU
anchors, with median absolute shifts of 2.95\,s, 1.46\,s, and 0.95\,s.

For HoloAssist, the released annotations were frozen before the
experiments, and the audit is a quality check of that frozen release
rather than a revision of it. After the freeze, we manually checked 399
utterances across eight test sessions, 9.75\% of the released
utterances, on all annotation layers except actions.
Figure~\ref{fig:annotation-holoassist-errors} reports the share of
audited items that the auditors changed on each layer. All reported
experiments use the released annotations as frozen; the corrections
identified in this audit are not applied to them, so every system is
scored against the same annotations.

\subsubsection{Release and Reproducibility Notes}

Every utterance, action/state, trigger link, and timing record has a
stable ID, and records refer to one another by these IDs rather than by
matching transcript text. To reproduce or extend the annotations, follow
the controlled values and boundary rules above, and leave uncertain cases
explicitly marked, for example with \texttt{AMBIGUOUS} validity or an
audit flag, rather than assigning a confident label.

\subsection{Annotation Examples and Boundary Cases}
\label{app:annotation_examples}

This section first lists the label distinctions that depend on
conversational context and then follows one intervention through all four
layers.

\subsubsection{Utterance-Label Boundary Cases}
\label{app:utterance_decisions}

A turn's role in the exchange determines its label, even when its wording
is the same. The rules below separate requests, uptake, and task feedback.

\paragraph{U1 versus U2.}
Use U2 only when the user proposes a concrete candidate that can be
affirmed or rejected; otherwise use U1. Thus \textit{Which spoon should I
use?} is U1, whereas \textit{Should I use this spoon?} is U2. A
declarative such as \textit{That should be enough} is U2 when it
functions as a bid for validation.

\paragraph{U2 versus U3.}
Use U3 when the object of the request is the instructor's recent words
themselves; a question about the task world is U2 even if it follows an
instruction. \textit{What do you mean by pinch?} is U3 with
\texttt{repair\_\allowbreak target=meaning}, and \textit{Thirty seconds,
you mean?} is U3 with \texttt{task\_\allowbreak detail}, whereas a delayed
check of remembered content, \textit{I think you said it was a teaspoon
earlier, right?}, is U2.

\paragraph{U4, U5, U9, and U10.}
\textit{Okay} or \textit{yeah} is U4 when it answers an instructor
question, U5 when it takes up a completed instructor turn, U10 when it
occurs while the instructor is still speaking, and U9 when it is
self-directed. An echo is U5 when the user closes it (\textit{Twelve
ounces? Okay.}) and U3 when the user waits for confirmation
(\textit{Twelve ounces?}).

\paragraph{I1, I2, I3, and I9.}
I1 supplies content requested explicitly or implicitly by the user; the
same content is I3 when volunteered as a required step. For a proposed
candidate, I2 accepts it and I9 rejects or replaces it. Thus
\textit{The teaspoon} answers \textit{Which spoon?} as I1, while
\textit{Yes} after \textit{This teaspoon?} is I2 and
\textit{No, use a teaspoon} after \textit{Should I use a spoon?} is I9.

\paragraph{I3 versus I4 versus I9.}
I3 is required guidance that presupposes no error; I4 is optional or
cautionary guidance with no existing error (\textit{Be careful not to add
too much salt}); I9 repairs an actual, proposed, imminent, or omitted
error (\textit{You added too much salt; remove some}). Modal wording such
as \textit{should}, \textit{can}, or \textit{might want to} does not
determine the label.

\paragraph{I10 and I11 versus Task Feedback.}
A receipt token such as \textit{okay} is I10 after a completed user
contribution and I11 during ongoing speech or action; if it evaluates
correctness, it is I2. For example, \textit{Okay} after the user answers
\textit{Five} is I10, while \textit{Yes} after \textit{Is that correct?}
is I2.

\subsubsection{End-to-End Four-Layer Example}
\label{app:worked_example}

\begin{table*}[t]
\centering
\small
\setlength{\tabcolsep}{5pt}
\renewcommand{\arraystretch}{1.12}
\begin{tabularx}{\textwidth}{@{}>{\raggedright\arraybackslash}p{0.28\textwidth}>{\raggedright\arraybackslash}p{0.2\textwidth}>{\raggedright\arraybackslash}X@{}}
\toprule
\textbf{Layer} & \textbf{Time (s)} & \textbf{Annotation} \\
\midrule
1. Utterance Taxonomy &
83.35--87.85 &
\textit{You should use the knife instead of the fork}; I9 Correction;
\texttt{content\_\allowbreak target=object\_\allowbreak tool}. \\
2. Action and State &
81.20--83.80; 85.54--87.83 &
\texttt{grab--fork}; validity \texttt{DEVIATION}; reviewed action
\texttt{a:356}. The subsequent \texttt{grab--knife} action follows the
correction. \\
3. Trigger Link &
action to utterance &
\texttt{USER\_\allowbreak ACTION\_\allowbreak STATE}; \texttt{ERROR}; high
confidence; no ambiguity flag; necessary. \\
4. Evidence Onset and Intervention Window &
ES 81.20; EO 82.84; LU 85.38 &
First sign of a wrong-utensil choice; earliest unambiguous fork evidence;
start of the compliant reach toward the knife (dialogue cutoff). \\
\bottomrule
\end{tabularx}
\caption{Complete four-layer example from session T2 (intervention
\texttt{T2\_\allowbreak R0005}).}
\label{tab:worked_example}
\end{table*}

Table~\ref{tab:worked_example} follows a wrong-utensil correction in
session T2 through the four layers. The first sign of the error appears at
81.20\,s (ES), and the fork becomes identifiable at 82.84\,s (EO). The
instructor starts the correction at 83.35\,s. At 85.38\,s, the user starts
a compliant reach toward the knife, which sets the dialogue cutoff and
therefore LU while the instructor is still speaking. The knife-grab span
itself starts only at 85.54\,s, when the grasped object becomes
identifiable, but LU marks the start of the reach. LU can precede the end
of the instructor turn because it concerns the usefulness of the
intervention's speech start time.

% Appendix C: Evaluation Protocol and Metrics.  Required packages: amsmath.

\section{Evaluation Protocol and Metrics}
\label{app:evaluation}

This appendix specifies the evaluation of Section~\ref{sec:evaluation}:
the teacher-forced protocol and its real-time execution, intervention
matching, the metrics, the LLM judges, and the statistical analysis. The
annotations it reads are defined in Appendix~\ref{app:annotation}.

\subsection{Evaluation Regime and Information Availability}
\label{app:eval-regime}

Every system is evaluated under a teacher-forced setting: the interaction
history it conditions on contains the human instructor's utterances,
released by the fixed rule of Appendix~\ref{app:eval-history}, while its own
utterances are recorded for scoring and do not re-enter its context. This
isolates each decision from earlier system errors and gives every system
the same instructor history. The user side of the history is what the
system itself perceives: the shared streaming-ASR transcript, or the user
audio for Gemini (Appendix~\ref{app:setup-common}).

\subsection{Teacher-Forced History Release Protocol}
\label{app:eval-history}

The $n$-th instructor utterance, spoken over $[t^{0}_n, t^{1}_n]$, enters
the history at
\begin{equation}
  \mathrm{release}_n=\max\bigl(t^{1}_n,\ t^{\mathrm{latest}}_n\bigr),
  \label{eq:eval-release}
\end{equation}
where $t^{\mathrm{latest}}_n$ is the annotated latest useful time of the
intervention it realizes (Appendix~\ref{app:timing_definitions}). An
utterance therefore becomes visible only after it has been fully spoken
and the opportunity it addresses has closed. An utterance without an
annotated window, such as an excluded intervention, enters at $t^{1}_n$,
and release times are made non-decreasing in transcript order, so the
history always follows the transcript order. A decision made at time $t$
sees utterance $n$ if and only if $\mathrm{release}_n<t$. How each
baseline inserts released utterances into its history is described in
Appendix~\ref{app:setup-baselines}.

\subsection{Real-Time Execution and Spoken Response Timing}
\label{app:eval-realtime}

Each system processes each session in real time. The locally served
systems (ProAssist, VideoLLM, JoyAI, and \Ours{}) run on one accelerator
each with batch size one: a model call starts when it is issued or when
the system's previous call has finished, whichever is later, and occupies
the accelerator for its measured duration, so queueing is part of every
latency. Perception, decision, generation, speech synthesis, and
sequential playback run on this clock; frames become available at their
capture time and final ASR transcripts at their recorded availability
time. Gemini is
accessed through the vendor-hosted Live API: its audio and video are
streamed at wall-clock speed, a response becomes audible when its first
audio arrives, and the time spent rebuilding a session
(Appendix~\ref{app:setup-baselines}) does not advance the replayed session timeline.

The evaluation reads every played system utterance $p$ with its audible
onset $t(p)$, its playback end $e(p)$, and the last instructor utterance
in the system's history when $p$ was decided; utterances that were queued
but never played are not candidates. Timing is read at the audible onset,
as the user experiences it.

\subsection{Intervention Event Construction and Matching}
\label{app:eval-matching}

\paragraph{Events and Windows.}
An intervention event is an instructor utterance of the test set whose
intervention status is \emph{necessary} or \emph{optional}; its side,
\emph{reactive} or \emph{proactive}, is defined in
Appendix~\ref{app:trigger_schema}. For event $n$ the annotation gives
the start $u_n$ and end $t^{\mathrm{usr}}_n$ of the triggering user turn
(reactive), the earliest sign $t^{\mathrm{early}}_n$ and the evidence
onset $t^{\mathrm{evid}}_n$ (proactive;
$t^{\mathrm{early}}_n=t^{\mathrm{evid}}_n$ when no earlier sign is
annotated), and the latest useful time $t^{\mathrm{latest}}_n$. The
\emph{intervention window} is
\begin{equation}
  W_n=
  \begin{cases}
    [t^{\mathrm{usr}}_n,\ t^{\mathrm{latest}}_n)   & \text{reactive},\\
    [t^{\mathrm{early}}_n,\ t^{\mathrm{latest}}_n) & \text{proactive}.
  \end{cases}
  \label{eq:eval-window}
\end{equation}
Candidates for event $n$ are the played utterances whose audible onset
falls in the \emph{matching window} $M_n=[L_n,U_n]$, which extends $W_n$
to the surrounding interval of adjacent reference utterances:
\begin{align}
  L_n&=
  \begin{cases}
    u_n & \text{reactive},\\
    \min(t^{1-}_n,\ t^{\mathrm{early}}_n) & \text{proactive},
  \end{cases}
  \label{eq:eval-match-lower}\\
  U_n&=\max\bigl(t^{\mathrm{latest}}_n,\ \min(t^{0+}_n,\ o^{+}_n)\bigr),
  \label{eq:eval-match-upper}
\end{align}
where $t^{1-}_n$ is the end of the previous instructor utterance,
$t^{0+}_n$ the start of the next one, and $o^{+}_n$ the opening time of
the next necessary event of the session, of either side ($u$ for a
reactive and $t^{\mathrm{early}}$ for a proactive event); the session
boundaries stand in when no such utterance or event exists. A candidate
is \emph{admissible} for $n$ if it was decided before the reference
entered the system's history, that is, if the last instructor utterance
in its history precedes utterance $n$.

\paragraph{Grading and Claiming.}
Each admissible candidate--event pair is graded once by the event judge
of the event's side (Appendix~\ref{app:eval-judge}) on the scale of
Table~\ref{tab:app-grades}. A pair is a \emph{content match} at grade
$\geq 4$; a grade-3 pair neither matches nor blocks a later match. Events
are processed in the order of the instructor utterances, optional events
included, and each
claims the earliest still-unclaimed matching candidate by audible onset,
regardless of timing. An utterance can satisfy at most one event. The
metrics are read from the necessary events.

\subsection{Reference Fidelity Metrics}
\label{app:eval-fidelity}

Similarity F1 and LLM-Judged F1 follow \citet{zhang2025proactive} and are
computed over all instructor utterances of the test set. A played
utterance $p$ may pair with a reference $r$ if
$t(p)-t^{0}(r)\in[-6,+3]$\,s and $p$ is admissible for $r$. For
Similarity F1, texts are encoded with the \texttt{all-mpnet-base-v2}
sentence encoder, pairs are assigned one-to-one by a cost that combines
the time offset with the cosine distance, and an assigned pair is correct
when the cosine similarity exceeds 0.5. For LLM-Judged F1, pairs are
pre-filtered to each reference's and each candidate's four most similar
partners, graded by the content judge (Appendix~\ref{app:eval-judge}), and
assigned one-to-one at grade $\geq 4$ with credit 1 (grade 5) or 0.75
(grade 4) so as to maximize total credit. Precision divides the correct
pairs (or credit) by the number of played utterances and recall by the
number of references.

\subsection{Timely Correctness and Response Latency}
\label{app:eval-correctness}

\textbf{Timely Correctness} assigns a temporal score to each necessary
intervention and averages over all of them,
\begin{equation}
  \begin{gathered}
  \frac{1}{N}\sum_{n=1}^{N}s_n,\\[2pt]
  s_n=
  \begin{cases}
    T_n(t_n), & \text{if intervention $n$ is correct},\\
    0,        & \text{otherwise},
  \end{cases}
  \end{gathered}
  \label{eq:timely-correctness}
\end{equation}
where $N$ is the number of necessary interventions of the side (reactive
or proactive) and $t_n$ is the audible onset of the claimed utterance.
Intervention $n$ is \emph{correct} when it claims an utterance
(Appendix~\ref{app:eval-matching}) whose audible onset falls in its
intervention window $W_n$; necessary interventions that are unmatched,
incorrectly matched, or delivered outside their valid timing windows
receive zero. The preferred response time is $t_{\mathrm{usr}}$ for a
reactive intervention and $t_{\mathrm{evid}}$ for a proactive
intervention. With the normalized positions
\begin{equation}
\begin{gathered}
u_E=\frac{t_{\mathrm{evid}}-t}{t_{\mathrm{evid}}-t_{\mathrm{early}}},\qquad
u_L=\frac{t-t_{\mathrm{evid}}}{t_{\mathrm{latest}}-t_{\mathrm{evid}}},\\[2pt]
u_R=\frac{t-t_{\mathrm{usr}}}{t_{\mathrm{latest}}-t_{\mathrm{usr}}},
\end{gathered}
\label{eq:tc-normalized-positions}
\end{equation}
the temporal function of a proactive intervention is
\begin{equation}
T_n^{\mathrm{pro}}(t)=
\begin{cases}
1-\lambda_E\,u_E^{2}, & t\in[t_{\mathrm{early}},t_{\mathrm{evid}}),\\[2pt]
\cos^{p_L}\!\bigl(\tfrac{\pi}{2}u_L\bigr), & t\in[t_{\mathrm{evid}},t_{\mathrm{latest}}),\\[2pt]
0, & \text{otherwise},
\end{cases}
\label{eq:tc-proactive}
\end{equation}
and that of a reactive intervention is
\begin{equation}
T_n^{\mathrm{rea}}(t)=
\begin{cases}
\cos^{p_L}\!\bigl(\tfrac{\pi}{2}u_R\bigr), & t\in[t_{\mathrm{usr}},t_{\mathrm{latest}}),\\[2pt]
0, & \text{otherwise}.
\end{cases}
\label{eq:tc-reactive}
\end{equation}
We use $(\lambda_E,p_L)=(0.7,0.5)$, calibrated in
Appendix~\ref{app:human-timing-study}. With these values, the proactive
score starts at $0.3$ at the earliest sign, peaks at the evidence onset,
and is zero at the latest useful time, and the exponent $p_L$ allows
moderately delayed responses to retain partial credit. If $t_{\mathrm{early}}=t_{\mathrm{evid}}$, the early
branch is empty.

\textbf{Correctness}, reported in Figure~\ref{fig:correctness-by-type},
is the share of necessary interventions that are correct, that is,
Eq.~(\ref{eq:timely-correctness}) with $T_n\equiv1$.

\paragraph{Response Latency.}
Response Latency is the median of $t_n-t_{\mathrm{usr}}$ over the
reactive necessary interventions whose claimed utterance begins at or
after the end of the user's turn. It is conditional on answering, so a
system that answers few turns can still have a low latency, and its
interval can be wide: Gemini's HoloAssist latency in
Table~\ref{tab:main-results} is a median over 11 content-matched reactive
responses from eight sessions.

\subsection{Intervention Precision and Turn-Taking}
\label{app:eval-discipline}

\paragraph{Intervention Precision.}
Let $P$ be the played non-empty utterances of a run and
$P^{\star}\subseteq P$ the utterances that are the claimed response of
some necessary \emph{or optional} event $n$ and whose audible onset lies
in $W_n$. Intervention Precision is $|P^{\star}|/|P|$. Utterances that
match no event, repeat a matched one, contradict the reference, or start
outside the window lower the score. Optional events count because
answering one constitutes a valid intervention, while leaving one
unanswered costs nothing.

\paragraph{Speech Overlap Ratio.}
With $\bar U$ the union of the session's user speech intervals, the same
for every system, and $S=\bigcup_{p\in P}[t(p),e(p))$, the ratio is
$|S\cap\bar U|/|\bar U|$. The user audio is replayed and does not react
to the system, so every overlap is attributed to the system.

\subsection{LLM Judges and Evaluation Context}
\label{app:eval-judge}

\paragraph{Configuration.}
Content agreement is the one part of the evaluation decided by a model;
admissibility, claiming, timing, overlap, and aggregation are
deterministic. One self-hosted Qwen3.8-27B model (bf16, temperature 0)
serves every judge call for every system and run. Each call grades one
candidate--reference pair from 1 to 5 with a short reason; only the grade
enters a metric, and an unparseable output counts as no match. The
complete judge prompts are provided in the released code repository.

\paragraph{Three Judges, One Scale.}
The \emph{reactive event judge}, the \emph{proactive event judge}, and the
\emph{content judge} share the grading scale of Table~\ref{tab:app-grades},
the context rules, and the output format. Each asks whether, after
hearing the candidate, the user would do and understand the same thing as
after the reference, where the user's need at that moment is read from
the trigger, the preceding dialogue, and the task knowledge rather than
from the reference's wording. For the reactive event judge the user's
turn sets the need; for the proactive event judge the observed action or
state sets it, so that, for example, approving an action the instructor
corrects is a contradiction. The content judge, used for LLM-Judged F1,
receives no trigger and grades every reference utterance by its
communicative function first, so a reference with no task content can be
missed but not contradicted. A candidate that differs from the reference
is credited when the task knowledge supports it and it leads to the same
next action or understanding.

\begin{table}[tb]
\centering
\small
\setlength{\tabcolsep}{3pt}
\renewcommand{\arraystretch}{1.08}
\begin{tabularx}{\columnwidth}{@{}c>{\raggedright\arraybackslash}X>{\raggedright\arraybackslash}p{0.22\columnwidth}@{}}
\toprule
\textbf{Grade} & \textbf{Meaning} & \textbf{Use} \\
\midrule
5 & The same guidance effect: the user would do and understand what the reference leads them to. & Match; F1 credit 1 \\
4 & The same effect with a non-critical difference, such as an extra supported detail or a supported alternative. & Match; F1 credit 0.75 \\
3 & Partly meets the need: something still needed is missing, or an unsupported claim could change what the user does. & No match; \emph{partial} \\
2 & Does not address this need: another turn, step, or problem, or narration where the reference guides. & No match; \emph{off-target} \\
1 & Works against this need: the user would do the opposite of what the reference directs. & No match; \emph{contradiction} \\
\bottomrule
\end{tabularx}
\caption{Grading scale shared by the three judges.}
\label{tab:app-grades}
\end{table}

\paragraph{Judge Context.}
The judges read the benchmark's ground-truth transcript, not the
streaming-ASR transcript or audio that the evaluated system perceived.
Every call contains the task knowledge text that every system received
(Appendix~\ref{app:setup-common}), the \emph{event context} (the benchmark
dialogue before the reference, with the event's own turn or observation
marked), the event, the reference, the \emph{candidate context}, and the
candidate; the content judge receives the same message without the event
block. The candidate context is the benchmark transcript cut at the
candidate's input cutoff, the start of the model call that produced it:
the user turns that had started by then, the instructor utterances
released into the system's history by then, and the annotated user action
in progress at that moment together with the last completed one. A user
turn that was still in progress at the cutoff is given in full, so the
candidate context can include the end of a turn that the system had not
yet heard; it approximates what the system had available rather than
reproducing it. It is kept separate from the event context, which may
extend past the cutoff, so that the candidate is interpreted in the
situation in which it was produced: a reply such as ``yes'' means what it
answered at that time, and a candidate produced before the event occurred
is off-target rather than a contradiction. A judge sees no timestamps,
trigger types, necessity labels, windows, or system identity.

\subsection{Aggregation and Statistical Analysis}
\label{app:eval-statistics}

Every metric is pooled over the test sessions of a dataset: event metrics
over events, Intervention Precision over played utterances, the Speech
Overlap Ratio over summed user speaking time, the F1 scores from pooled
precision and recall, and Response Latency as the pooled median. Every
reported value carries a 95\% session-cluster bootstrap interval (10,000
resamples of sessions; tables print the half-width of the 2.5--97.5
percentile interval).

ProAssist, VideoLLM, JoyAI, and \Ours{} were each run twice under the
same configuration and the reported values are the mean of the two runs;
Gemini and the two ablation variants of Table~\ref{tab:ablation} were run
once. A difference between two systems or variants is
computed on identical session sets as the pooled difference with a paired
session-cluster bootstrap interval.

\section{FLUID-Agent Implementation}
\label{app:egoguide}

This appendix describes \Ours{} (Section~\ref{sec:framework},
Figure~\ref{fig:framework}) as run in every experiment, with one
configuration on both datasets. A single frozen Qwen3.6-35B-A3B
vision-language model (FP8 weights, vLLM, temperature 0) serves every
online role through task-specific prompting: it perceives the scene,
judges the task state, interprets and answers user turns, and generates
spoken intervention utterances. Deterministic code around the model maintains the
shared task memory, schedules the calls, evaluates the evidence
conditions under which an intervention is opened, and manages
turn-taking. The model therefore supplies the evidence, and the agentic
structure decides what that evidence licenses. 

\paragraph{Session Loop.}
When a visual change is detected, the perception module analyzes the video and adds an observation report to the task memory. The \emph{Situation Tracker}
(the task-state tracking part of the Task Memory block in
Figure~\ref{fig:framework}) reassesses the task state when the observed
activity or the dialogue has changed, and the intervention conditions are re-evaluated after each
judgment. In parallel, each completed user turn is handled by one call
that interprets it and, when a reply is due, answers in the same
model call. Under the teacher-forced protocol
(Appendix~\ref{app:eval-history}), released instructor utterances are
visible to perception, to the Tracker as testimony of what the user was
told, and to utterance understanding; the system's own outputs do not
enter its prompts.

\subsection{Perception and Task-State Tracking}
\label{app:ego-perception}

\paragraph{Task Knowledge Compilation.}
\label{app:ego-plan}
An offline compiler (Qwen3.8-27B, distinct from the online backbone)
reads the same task knowledge that every baseline receives
(Appendix~\ref{app:setup-common}) and produces an action-level plan: each
step becomes a few imperative actions with the tools and destinations the
step names, its stated quantities, an observable completion check where
the knowledge states one, and the step's prerequisite relations.
Task-specific procedural knowledge reaches the online model only through
this plan, which supplies the checks the perception module answers, the
reference for assessing the user's progress through the task plan and
detecting deviations, and the content of intervention utterances. The dialogue and the visual observations
supply the task context of the moment.

\paragraph{Perception Calls.}
A novelty signal computed from DINOv2 features of the incoming frames
triggers a perception call at a novelty peak, during sustained activity,
or after a fixed interval. The perception call receives recent keyframes
together with the dialogue associated with the current step and its
completion checks, and reports the user's hand activity, the current
action, visible state changes, the check values, and per-step progress,
marking whether each value was seen or inferred from the dialogue.

\paragraph{Task Memory.}
\label{app:ego-memory}
Perception reports are deterministically integrated into a belief state
that tracks task progress, committed completion checks, and recent
evidence for each step. A check value is committed when consecutive
perception calls agree, a committed completion is retained while the
check is out of view, and the memory records per step how often the
action associated with that step was observed. The Tracker, the
intervention conditions, and utterance understanding read this belief
state rather than individual reports.

\paragraph{Situation Tracker.}
The Tracker is a text-only call. It runs after a perception update has
been integrated into the belief state when the observed activity has
changed, when a user turn or a released instructor utterance has arrived
since its last judgment, when a pending intervention candidate awaits
confirmation, or otherwise after a bounded interval, with a minimum
spacing between calls. From the plan, its
previous judgment, the step evidence of the belief state, and the recent
reports interleaved with the dialogue, it returns the current step and
action, each step's state with its basis (observed, moved past, or the
user's testimony), at most one deviation with whether it matters, and
wait and stuck states. It weighs evidence conservatively: the user's
behavior rather than the plan order defines the user's position in the
task plan, a deviation
requires agreeing consecutive reports or one report plus the user's own
words, and instructor utterances count as testimony of what the user was
told, not as evidence of what was done. A consequential deviation is
accepted only when consecutive judgments agree, and a stuck state only
when it persists; the resulting position in the task plan sets the checks
of the next perception call.

\subsection{Proactive and Reactive Intervention Decisions}
\label{app:ego-intervention}

\paragraph{Evidence-Conditioned Proactive Decision.}
After each accepted judgment, a fixed set of intervention conditions is
checked against the current belief state. When a condition is satisfied,
the agent creates a typed \emph{intervention candidate} (typed by the
trigger types of Section~\ref{sec:interaction-structure}) grounded in the
current evidence, which may subsequently be spoken, held, or discarded as
that evidence develops. \emph{Correction} opens on an accepted
consequential deviation and states the contrast between what the user is
doing and what the action calls for. \emph{Progress coaching} opens at
the onset of an action with a stated amount, manner, or critical
parameter while the user is manipulating task objects and no deviation
is open.
\emph{Completion and hand-off} opens when a step is reliably done: the
final step receives a closing confirmation, and a completed non-final
step is followed by the next step's instruction unless that step has
already been given, by the system or by a released instructor utterance,
or has already started or passed. \emph{Stall help} opens on a persisting
stuck judgment. A ledger makes sure that the same evidence is voiced only
once per visit to a step, and a pending candidate is discarded when its
evidence is withdrawn or its window closes. The content of an intervention
utterance is assembled from the plan (and, for corrections, from the Tracker's
deviation description, which states what the user is doing and what the
action calls for), and one short model call rephrases it as a spoken sentence.

\paragraph{Taxonomy-Guided Reactive Decision.}
\label{app:ego-utterance}
Each final ASR transcript with speech content is handled by one model
call; final transcripts without words and very short acknowledgment or
filler transcripts are filtered without invoking the model. The call labels the turn with one of the
user-utterance categories of Table~\ref{tab:user_taxonomy} and, for
labels that call for a reply, continues in the same model call with the
answer. The label places the turn in one of three response tiers:
requests for information, verification, or clarification and trouble
reports are answered. Narration and answers to a pending question are
voiced only when the user's turn states something that can be checked
against the task context and the model produces a brief confirmation or
correction. Plain narration receives no reply, and the remaining labels
are silent. A reply is one or
two short spoken sentences grounded in the plan, the latest situation
judgment, and the recent dialogue, and a failed call produces no
response.

\paragraph{Turn-Taking and Speech.}
\label{app:ego-speech}
While the system speaks, incoming partial ASR transcripts are checked: a
filler-only partial transcript lets it continue, and a question cue or a
label requiring a reply causes the system to defer to the user. Deferring
stops the current sentence, holds queued proactive responses until the
user has finished (dropping those whose step is no longer current), and
cancels unstarted answers.
Sentences are synthesized by the shared speech server
(Appendix~\ref{app:setup-common}) and played one after another while
perception continues.

\subsection{Ablation Implementation}
\label{app:ego-ablation}

The two variants of Section~\ref{sec:ablation} run the final system with
one switch each. Each replaces one decision mechanism, together with the
control logic attached to it, by a direct decision of the same model on
the evidence that mechanism used, while the backbone, perception, task
memory, and Situation Tracker stay those of the final system. The
contrasts therefore show what the taxonomy-guided reactive decision and
the evidence-conditioned proactive decision contribute to the system's
selectivity. They do not isolate the contribution of any single rule,
since each variant also differs from the final system in the control
logic noted below.

\paragraph{Direct Reactive.}
This variant replaces the taxonomy-guided utterance understanding (the
category label, the response tiers, and the filler guard) by one direct
decision of the model on the same turn context. The prompt keeps the
reply rules without reference to the taxonomy, states no preference for
silence, and asks the model to decide whether the turn needs a spoken
reply now and, if so, to write it in the same call. Turn-taking on
partial hypotheses and the proactive side are unchanged.

\paragraph{Direct Proactive.}
This variant replaces the evidence-conditioned proactive decision (the
intervention conditions, the candidate ledger, and the templated content with
its phrasing step) by one direct decision of the model every 2\,s of
session time, skipped while a previous decision is still being computed.
The decision reads the same evidence products (the plan, the latest
Tracker judgment, the step evidence, and the recent reports with the
dialogue) together with the utterances the user has recently heard, and
returns either a response text, spoken as written, or nothing. Proactive
utterances are thus generated on a fixed schedule rather than on
evidence-triggered candidates, and a newer utterance supersedes an older
one that has not started playing; replies
to user turns and speech already playing are unaffected. The reactive
side is unchanged.

\paragraph{Ablation Diagnostics.}
Table~\ref{tab:ablation} classifies each played utterance as reactive (a
reply to a user turn) or proactive (an utterance the system initiated), as
the system
labels its own output. \emph{Reactive} and \emph{proactive Intervention
Precision} restrict Intervention Precision
(Appendix~\ref{app:eval-discipline}) to the utterances of that side, and
\emph{utterances per minute} count them per minute of session time. The
\emph{response rate on unanswered turns} measures the proportion of user
turns that did not require an instructor response but nevertheless
received a system response. A user turn is considered not to require a
response if it is not linked as the trigger of any necessary or optional
reactive intervention. Each turn is paired with its most-overlapping final
ASR transcript; turns without a corresponding final transcript are
excluded. The \emph{repetition rate} is the share of proactive
utterances whose text, after case and punctuation normalization, repeats
an earlier proactive utterance of the same session verbatim.

\section{Experimental Setup and Baselines}
\label{app:setup}

This appendix gives the conditions shared by the systems of
Section~\ref{sec:setup} and the adaptation of each baseline. \Ours{} is
described in Appendix~\ref{app:egoguide} and the evidence-ladder
conditions in Appendix~\ref{app:ladder}.

\subsection{Common Experimental Setup}
\label{app:setup-common}

\paragraph{Systems.}
We evaluate \Ours{} and the four baselines of Section~\ref{sec:setup},
each with its released checkpoint or hosted model. To be compatible with
the teacher-forced protocol, a system must allow instructor utterances to
be inserted into its history at a specified time without its own earlier
outputs; Appendix~\ref{app:setup-baselines} states how each baseline
realizes the release rule of Appendix~\ref{app:eval-history}. All systems
run each session in real time on the clock of
Appendix~\ref{app:eval-realtime}.

\paragraph{Shared Inputs.}
\emph{User audio and transcript:} the three open-weight baselines and
\Ours{} hear the user through one shared causal streaming-ASR stream
(NVIDIA Nemotron streaming speech recognition, 0.6B parameters), produced
once per session from the user-only audio track
(Appendix~\ref{app:multimodal-preprocessing}) and replayed for every
system with each final transcript's recorded availability time, so
recognition errors are shared; Gemini hears the same audio natively. \emph{Video:}
frames are decoded once per session by source timestamp; ProAssist and
VideoLLM read the latest frame at 2\,fps, JoyAI samples at its native
1\,s ticks, Gemini receives 1\,fps (the Live API maximum), and the perception
trigger of \Ours{} runs at 8\,fps. \emph{Task knowledge:} every system
receives the same structured task knowledge of a session, extracted from
egocentric demonstration video by a multi-round vision-language pipeline:
one text per task for WTaG, whose sessions follow a common recipe, and
one text per session for HoloAssist, whose procedures vary across
sessions. The text lists the steps with an instruction sentence and
prerequisite relations and, for WTaG, the objects of the task and a
completion criterion per step. It is rendered once and provided to each
system in the input format its model expects; the plan compiler of
\Ours{} (Appendix~\ref{app:ego-plan}) and the LLM judges
(Appendix~\ref{app:eval-judge}) read the same text. \emph{Speech output:}
the open-weight baselines and \Ours{} speak through the same Kokoro
text-to-speech server, sentence by sentence; Gemini produces speech
natively, and the API's output transcription is the scored text.

\paragraph{Models and Hardware.}
Table~\ref{tab:app-hardware} lists the backbone, hardware, and decoding
of each system. Accelerators differ across systems and Gemini runs on
vendor infrastructure, so Response Latency reflects end-to-end performance
under each system's actual deployment configuration and should not be
interpreted as a controlled comparison of model inference speed. Gemini
was accessed through its API in September 2026. Besides the streaming-ASR
and text-to-speech servers above, \Ours{} uses DINOv2 perception-trigger
features computed on the CPU of the machine running \Ours{}, as well as
offline plan compilation with Qwen3.8-27B (Appendix~\ref{app:ego-plan}).
For evaluation, one Qwen3.8-27B judge is shared across all systems
(Appendix~\ref{app:eval-judge}). The number of runs per system is given
in Appendix~\ref{app:eval-statistics}.

\begin{table}[tb]
\centering
\small
\setlength{\tabcolsep}{3pt}
\renewcommand{\arraystretch}{1.08}
\begin{tabularx}{\columnwidth}{@{}l>{\raggedright\arraybackslash}X>{\raggedright\arraybackslash}p{0.19\columnwidth}>{\raggedright\arraybackslash}p{0.25\columnwidth}@{}}
\toprule
\textbf{System} & \textbf{Backbone} & \textbf{Hardware} & \textbf{Decoding} \\
\midrule
ProAssist & Llama-3.1-8B + SigLIP & A40 & greedy; speak gate $\theta=0.2$ \\
VideoLLM & Llama-3-8B-Instruct + SigLIP & A40 & greedy; upstream keep-watching threshold \\
JoyAI & JoyAI-VL-Interaction & RTX PRO 6000 & official sampling, fixed seed \\
Gemini & Gemini 3.8 Live (Live API) & hosted & vendor defaults \\
\Ours{} & Qwen3.6-35B-A3B (FP8), one instance & RTX PRO 6000 & temperature 0 \\
\bottomrule
\end{tabularx}
\caption{Models, hardware, and decoding.}
\label{tab:app-hardware}
\end{table}

\subsection{Baseline Configurations and Adaptations}
\label{app:setup-baselines}

\paragraph{Shared Guidance Instructions.}
Every baseline receives the same task-guidance objectives in the slot its
training format uses for instructions: a role description (the camera is
the user's first-person view; speak when it helps, by confirming a
completed step and giving the next one, correcting a visible mistake,
helping when the user is stuck, or answering a question; otherwise stay
silent), the task-knowledge text, delivery rules (one short spoken
sentence, no preamble, no repetition, no scene description), and the
instruction to speak only when supported by concrete visual or auditory
evidence. The
wording follows each model's native prompt format, and the prompts are
provided in the released code repository.

\paragraph{ProAssist.}
ProAssist receives frames at its native 2\,fps with the official
preprocessing. At every frame the speak gate compares the probability of the
end-of-turn (not-talk) token with the threshold $\theta=0.2$, which lies
within the range of the model's own ablation; the model stays silent when
the probability exceeds the threshold and otherwise generates with the
official greedy decoding loop. The dialogue opens in the checkpoint's training form
with the shared instructions and the knowledge text; final ASR transcripts
enter as user turns and released instructor utterances as assistant turns.

\paragraph{VideoLLM.}
VideoLLM also receives frames at 2\,fps. Its streaming loop is
re-implemented in-process: at each frame the probability of the
keep-watching token is compared with the upstream default threshold, and
a response is decoded greedily when that probability falls below the
threshold. The system prompt is the
checkpoint's own, and the shared instructions with the knowledge text
form the first user turn, the trained instruction slot. Final ASR
transcripts enter as user turns and released instructor utterances as assistant turns, and
the context is rebuilt without the system's own outputs.

\paragraph{JoyAI.}
The released checkpoint is served by a bare vLLM server, and the live
adapter (ticks, action tokens, frame window) is re-implemented in our
evaluation code, so the evaluated system is the released checkpoint with our
real-time reconstruction of its adapter. At every 1\,s tick the model
receives the last 140 ticks in native format (frames, final ASR
transcripts, and released instructor utterances at their ticks), with
earlier ticks kept only as their text turns and the history rebuilt
without its own outputs, and emits either a silence token or a response
token followed by text. Decoding uses the official adapter's sampling
with a fixed per-request seed.

\paragraph{Gemini.}
The vendor-hosted Live model hears the user audio natively, watches the
video at 1\,fps, and decides itself when to speak. Its system instruction
carries the shared instructions adapted to native audio and a note that
the history it starts from is the real record of the session, with the
instructor's utterances in its own role. The teacher-forced history is
realized by \emph{rebuilding}: whenever an instructor utterance is
released, the current Live session is closed and a new one is opened
whose initial content is the recorded interaction so far (the user's
audio, frames at a reduced rate for the earlier part of the session, and
the released instructor utterances as model turns), after which real-time
streaming resumes. The model's own earlier outputs are not part of a
rebuilt history, and output produced while a history is being seeded is
discarded.

\section{Evidence Ladder Experimental Design}
\label{app:ladder}

The evidence ladder (Section~\ref{sec:evidence-bottlenecks},
Table~\ref{tab:evidence-ladder}) was run on the 43 WTaG test sessions
with JoyAI and Gemini. Its levels change the conditions under which a
system intervenes at the annotated necessary interventions and the
information it receives about them, while what to say is left to the
system at every level; no level gives the reference response. This
appendix defines the levels and then describes how they are implemented
for each system.

\subsection{Ladder Conditions}
\label{app:ladder-conditions}

\paragraph{Triggers.}
The triggers are the necessary WTaG interventions at their annotated
onset: the end of the user's turn for interventions whose trigger
includes a user utterance, and the evidence onset for the others.
Annotated relations from the same source within 1\,s of each other are
merged into one intervention. Interventions whose trigger times fall in
the same one-second decision step form one trigger, timed at the latest
of their trigger times; the same decision-step rule is used for JoyAI and
Gemini. Table~\ref{tab:app-ladder-levels} defines the four levels.

\begin{table}[tb]
\centering
\small
\setlength{\tabcolsep}{3pt}
\renewcommand{\arraystretch}{1.08}
\begin{tabularx}{\columnwidth}{@{}l>{\raggedright\arraybackslash}X>{\raggedright\arraybackslash}X@{}}
\toprule
\textbf{Level} & \textbf{Intervention decision} & \textbf{Information at the trigger} \\
\midrule
L0 & Native: the system decides itself whether and when to speak. & None. \\
L1 & Enforced: at each trigger the system is made to attempt one response; its speech elsewhere is suppressed. & That an intervention is required now. \\
L2 & As L1. & L1 plus the trigger type of each intervention, with a one-line gloss (Section~\ref{sec:interaction-structure}). \\
L3 & As L1. & L2 plus a short observed-fact sentence for the interventions whose trigger includes annotated action or state evidence and for which such a sentence exists; the other interventions keep the L2 note. \\
\bottomrule
\end{tabularx}
\caption{Conditions of the evidence ladder.}
\label{tab:app-ladder-levels}
\end{table}

At a trigger, the note is one sentence in the request (``Intervene now.
Evidence requiring an instructor intervention is available. Type:
\emph{not provided} $|$ \emph{type (gloss)}.''), to which L3 appends
``Observed (video): \emph{fact}.''; it is not kept in later history. The
observed-fact sentences are short descriptions derived from the action
and state annotations (Appendix~\ref{app:annotation}). An enforced
response attempt can fail to produce a playable utterance (an empty
generation, or a response not received within the API's time limit), and
such triggers count as no intervention. For JoyAI, a trigger that
includes a user utterance also receives the annotated transcript of that
utterance in place of the final ASR transcript, which often arrives
after the annotated end of the turn; Gemini hears the audio itself.

\paragraph{Reading the Ladder.}
L1 changes the intervention regime as a whole, since the system is told
the annotated onset, made to respond there, silenced elsewhere, and, for
JoyAI, given the annotated transcript of the triggering turn and the
ladder prompt.
$\Delta_{\mathrm{onset}}$ therefore measures the combined effect of these
changes, not the contribution of timing information alone.
$\Delta_{\mathrm{type}}$ and $\Delta_{\mathrm{fact}}$ add information to
the note under the same regime; because the levels are separate runs,
their differences also contain run-to-run variation. L3 adds observed
facts only for interventions with action or state evidence, so its effect
concentrates on proactive guidance
(Section~\ref{sec:evidence-bottlenecks}). The values at L1--L3 describe
the systems under enforced intervention and are not estimates of their
native capability.

\subsection{Implementation}
\label{app:ladder-mechanism}

\paragraph{JoyAI.}
At a trigger tick the note is appended to the request and the decoder is
made to speak by pre-filling the assistant turn with the response action
token, so that the model writes the intervention; at other ticks the
request is made as usual, and a decision to speak is recorded but not
played. The decision time of an enforced intervention is the start of
the inference at the trigger tick. L1--L3 share one system prompt, which
describes the note and differs in wording from the main-experiment prompt
(Table~\ref{tab:main-results}); L0 is the main-experiment run
(Table~\ref{tab:main-results}).

\paragraph{Gemini.}
The Live API generates at activity boundaries, so each trigger is
realized as one fresh Live session seeded with the recorded interaction
up to the last released instructor utterance (the rebuild of
Appendix~\ref{app:setup-baselines}). The audio and video up to the
trigger are then streamed in real time inside one manual activity with
automatic activity detection disabled, and the note is sent as text
followed by the end of the activity, which marks the session's only turn
boundary and triggers generation. Output produced before the note is discarded.
The system instruction is the same across L1--L3, and the
main-experiment run (Table~\ref{tab:main-results}) serves as L0.

% Appendix G: Human Timing Study.  Required packages: amsmath, booktabs.

\section{Human Timing Study}
\label{app:human-timing-study}

This study tests the annotated timing windows and calibrates the graded
credit used by Timely Correctness. Three raters judged the timing of WTaG
interventions whose content was held fixed while only the speech start
time was varied, so that the ratings reflect timing rather than wording.
We first describe the study events, the candidate moments, and the
rating procedure, then compare the annotated windows with the human
judgments of acceptable timing, and finally compare temporal scoring
functions and select their parameters with leave-one-session-out (LOSO)
cross-validation. Throughout, the anchors are written as in
Appendix~\ref{app:timing_definitions}: ES, EO, LU, and R correspond to
$t_{\mathrm{early}}$, $t_{\mathrm{evid}}$, $t_{\mathrm{latest}}$, and
$t_{\mathrm{usr}}$ of the main text and Appendix~\ref{app:eval-correctness}.

\subsection{Study Events}
\label{app:timing-event-selection}

The formal study covers 79 WTaG events from 31 sessions, and a separate
set of 12 events was used for calibration. Formal events required
complete source media, a reference instructor utterance, and the anchors
needed for the reactive or proactive window. We included the 12 non-Other
instructor categories (Table~\ref{tab:instructor_taxonomy}) and excluded
reference utterances that overlap user speech, since these would confound
the speech start time with simultaneous speech.

Events were sampled with category quotas to cover the intervention types,
with at most three events per session and none from the calibration
sessions; the quotas are not intended to reflect corpus prevalence. The
formal set
contains 26 coffee, 28 mug-cake, and 25 pinwheel events: 43 reactive and
36 proactive, 46 necessary and 33 optional.

The calibration set contains one event from each of 12 distinct sessions,
four per task. It is balanced between reactive and proactive, necessary
and optional, and three anticipated difficulty levels, covers 11 of the
12 non-Other categories, and uses the same candidate-moment construction
as the formal study.

\subsection{Candidate-Moment Construction}
\label{app:timing-moment-sampling}

Each event has six candidate moments spanning early, valid, and late
responses. Two are fixed: the preferred time, which is the end of the
triggering user turn (R) for reactive events and the evidence onset (EO)
for proactive events, and the speech start time of the original
instructor utterance. The other four are sampled relative to the event's
anchors. The early moment precedes EO for proactive events and falls
within the triggering user turn for reactive events. Two moments cover
the first and second halves of the interval from the preferred time to
the latest useful time (LU). The last slot is sampled after LU with
probability 0.5, from a heavy-tailed distribution, and otherwise uniformly
between the earliest candidate position (ES for proactive events, the
start of the triggering user turn for reactive events) and LU.
Moments must be at least 0.1\,s apart; near-duplicate fixed moments are
merged and the open slot is resampled. Sampling therefore covers
positions relative to the anchors rather than uniformly across the video.

\subsection{Rating Protocol and Calibration}
\label{app:timing-rating-procedure}

Raters viewed the source context up to each candidate moment and then
judged the timing of the same synthesized reference utterance on an
integer scale from 0 (definitely should not intervene now) to 10 (exactly
the desired moment); they were asked to judge timing rather than wording.
The rubric marked 5 as borderline and 6 as acceptable but improvable, and
\texttt{Cannot judge} responses were treated as missing. All three raters
were authors. The roles of candidate moments (early, preferred, or late),
annotated timing boundaries, and original reference start-time labels were
hidden from raters.

All raters completed calibration before the formal study. The 12
calibration events produced 216 records, including one \texttt{Cannot
judge}. Calibration familiarized the raters with the interface and the
rubric; its ratings were excluded from the primary reference, MAE, and
parameter selection, and were not used to exclude or reweight raters.

The formal phase produced 1,422 rating rows over 474 candidate moments,
of which 1,415 are valid ratings and seven are \texttt{Cannot judge}. The
human reference for each moment is the mean available rating divided by
ten.

\subsection{Validation of the Annotated Timing Windows}
\label{app:timing-window-validation}

Before calibrating graded credit, we tested whether the annotated valid
windows agree with binary human judgments of acceptable timing. For
candidate $j$ of event $i$, let $\bar r_{ij}$ be the mean of the valid
rater scores on the 0--10 scale. Following the rubric's threshold for
acceptable timing, human acceptability is
\begin{equation}
A_{ij}=\mathbf{1}\bigl[\bar r_{ij}\geq 6\bigr].
\label{eq:timing-human-acceptability}
\end{equation}
We compare two binary timing rules:
\begin{align}
B^{\mathrm{win}}_{ij}&=\mathbf{1}\bigl[t_{ij}\in W_i\bigr],
\label{eq:timing-binary-window}\\
B^{[-6,+3]}_{ij}&=\mathbf{1}\bigl[-6\leq t_{ij}-t_{\mathrm{ref},i}\leq 3\bigr],
\label{eq:timing-binary-fixed}
\end{align}
where $W_i$ is the annotated valid window of event $i$
($[\mathrm{R},\mathrm{LU})$ for reactive and $[\mathrm{ES},\mathrm{LU})$
for proactive events) and $t_{\mathrm{ref},i}$ is the speech start time
of the original instructor utterance, in seconds. Each rule is treated as
a classifier of $A_{ij}$, and its agreement with the human judgments is
measured by Macro-F1, the unweighted mean of the F1 scores of the
acceptable and the unacceptable class:
\begin{equation}
\begin{gathered}
\mathrm{Macro\text{-}F1}=\tfrac{1}{2}\bigl(F_1^{+}+F_1^{-}\bigr),\\[2pt]
F_1^{c}=\frac{2\,\mathrm{TP}_c}{2\,\mathrm{TP}_c+\mathrm{FP}_c+\mathrm{FN}_c},
\end{gathered}
\label{eq:timing-macro-f1}
\end{equation}
where $F_1^{+}$ takes the acceptable moments ($A_{ij}=1$) as the positive
class and $F_1^{-}$ the unacceptable moments ($A_{ij}=0$), with
$\mathrm{TP}_c$, $\mathrm{FP}_c$, and $\mathrm{FN}_c$ counted for class
$c$. Unacceptable moments are the minority (146 of the 474 candidate
moments), so the macro average weights the two classes equally.
Confidence intervals and paired differences use 5,000 session-cluster
bootstrap draws, each resampling the 31 formal sessions with replacement
and keeping every candidate moment of a sampled session.

\begin{figure}[tb]
\centering
\includegraphics[width=\linewidth]{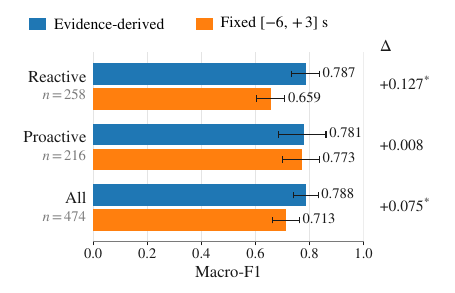}
\caption{\textbf{Agreement with human acceptability.} Macro-F1 of the
evidence-derived windows (blue) and the fixed $[-6,+3]$-second windows
(orange) against human acceptability (mean rating of at least 6), per
guidance side and pooled. Whiskers show 95\% session-cluster bootstrap
confidence intervals; $\Delta$ is the paired difference (evidence minus
fixed), starred when its 95\% interval excludes zero; $n$ counts
candidate moments.}
\label{fig:timing-window-f1}
\end{figure}

Figure~\ref{fig:timing-window-f1} reports the results. Over all 474
moments, the evidence-derived window reaches a Macro-F1 of $0.788$ (95\%
CI $[0.741,0.832]$) against $0.713$ ($[0.663,0.762]$) for the fixed
window, a paired difference of $+0.075$ ($[0.002,0.149]$). The gap comes
from reactive events, $0.787$ against $0.659$, a paired difference of
$+0.127$ ($[0.040,0.220]$): the fixed window accepts 61 unacceptable
reactive moments, 39 of them while the user is still speaking, whereas
the evidence window opens only after the user's turn ends. On proactive
events the two windows perform comparably, $0.781$ ($[0.686,0.861]$)
against $0.773$ ($[0.701,0.836]$), a difference of $+0.008$
($[-0.099,0.116]$); the fixed window rejects 16 acceptable moments that
lie inside the evidence window, 5 of them earlier than 6\,s before the
reference onset and 11 later than 3\,s after it.

This comparison tests the window boundaries. The following analyses test
how much credit Timely Correctness should assign within a valid window.

\subsection{Temporal Scoring Hypotheses}
\label{app:timing-hypotheses}

We compared three families of graded timing scores, each assigning zero
outside the valid window. They are written in terms of the anchors ES,
EO, R, and LU of Appendix~\ref{app:timing_definitions} and the normalized
positions
\begin{equation}
\begin{gathered}
u_E=\frac{\mathrm{EO}-t}{\mathrm{EO}-\mathrm{ES}},\qquad
u_L=\frac{t-\mathrm{EO}}{\mathrm{LU}-\mathrm{EO}},\\[2pt]
u_R=\frac{t-\mathrm{R}}{\mathrm{LU}-\mathrm{R}},
\end{gathered}
\label{eq:timing-normalized-positions}
\end{equation}
where $u_E$ falls from 1 to 0 over $[\mathrm{ES},\mathrm{EO})$, while
$u_L$ and $u_R$ rise from 0 to 1 over $[\mathrm{EO},\mathrm{LU})$ and
$[\mathrm{R},\mathrm{LU})$, respectively.

\paragraph{Hypothesis 1: Normalized Linear Position.}
Under this hypothesis, credit changes at a constant rate with the
normalized position in the event-specific window:
\begin{align}
T_{\mathrm{lin}}^{\mathrm{pro}}(t)
&=
\begin{cases}
1-u_E, & t\in[\mathrm{ES},\mathrm{EO}),\\
1-u_L, & t\in[\mathrm{EO},\mathrm{LU}),
\end{cases}
\label{eq:timing-linear-proactive}\\
T_{\mathrm{lin}}^{\mathrm{rea}}(t)
&=1-u_R, \qquad t\in[\mathrm{R},\mathrm{LU}).
\label{eq:timing-linear-reactive}
\end{align}

\paragraph{Hypothesis 2: Base-2 Squared Exponential.}
This hypothesis uses absolute elapsed seconds rather than the normalized
window position:
\begin{align}
T_{\mathrm{exp}}^{\mathrm{pro}}(t)
&=
\begin{cases}
2^{-((t-\mathrm{EO})/\tau_E)^2}, & t\in[\mathrm{ES},\mathrm{EO}),\\[2pt]
2^{-((t-\mathrm{EO})/\tau_L)^2}, & t\in[\mathrm{EO},\mathrm{LU}),
\end{cases}
\label{eq:timing-exponential-proactive}\\
T_{\mathrm{exp}}^{\mathrm{rea}}(t)
&=2^{-((t-\mathrm{R})/\tau_R)^2}, \qquad t\in[\mathrm{R},\mathrm{LU}),
\label{eq:timing-exponential-reactive}
\end{align}
with the WTaG corpus-level scales $\tau_E = 1.10\,\mathrm{s}$,
$\tau_L = 2.80\,\mathrm{s}$, and $\tau_R = 1.22\,\mathrm{s}$.

\paragraph{Hypothesis 3: Quadratic Onset and Cosine-Power Decay.}
This family controls the proactive early discount and the shape of the
late decline separately:
\begin{align}
T_{\mathrm{cos}}^{\mathrm{pro}}(t)
&=
\begin{cases}
1-\lambda_E\,u_E^{2}, & t\in[\mathrm{ES},\mathrm{EO}),\\[2pt]
\cos^{p_L}\!\bigl(\tfrac{\pi}{2}u_L\bigr), & t\in[\mathrm{EO},\mathrm{LU}),
\end{cases}
\label{eq:timing-cosine-proactive}\\
T_{\mathrm{cos}}^{\mathrm{rea}}(t)
&=\cos^{p_L}\!\bigl(\tfrac{\pi}{2}u_R\bigr), \qquad t\in[\mathrm{R},\mathrm{LU}).
\label{eq:timing-cosine-reactive}
\end{align}
Here $\lambda_E$ sets the proactive discount at ES, where the credit is
$1-\lambda_E$, and $p_L$ sets the decline after the preferred time. The
settings $(\lambda_E,p_L)=(0.2,4)$ and $(0.4,1)$ are reported as reference
points within this family, not as separate models.

As diagnostic baselines, we also score the two binary rules of
Eqs.~(\ref{eq:timing-binary-window}) and (\ref{eq:timing-binary-fixed})
as step functions, which compares graded credit with binary timing
acceptability.

\subsection{Selection Criterion and Hyperparameter Search}
\label{app:timing-selection}

Scoring functions are selected by event-level mean absolute error (MAE)
against the normalized human ratings. For a temporal function $T$,
\begin{equation}
  \operatorname{MAE}(T)
  =\frac{1}{|\mathcal{E}|}
   \sum_{i\in\mathcal{E}}
   \frac{1}{|\mathcal{M}_i|}
   \sum_{j\in\mathcal{M}_i}
   \left|T(t_{ij})-\bar y_{ij}\right|,
  \label{eq:timing-event-mae}
\end{equation}
where $\bar y_{ij}$ is the mean valid human rating divided by ten and
$\mathcal{M}_i$ contains the valid candidate moments of event $i$. Errors
are thus averaged within each event and then across events, giving events
equal weight. Mean within-event Kendall $\tau_b$ serves as a secondary
measure of ordering; MAE is primary because Timely Correctness uses the
amount of credit. Uncertainty intervals come from a session-cluster
bootstrap in which each draw resamples sessions and keeps every event and
candidate moment of each sampled session, and differences between
functions are paired within the same draw.

For the cosine family, we search
\begin{equation}
\begin{gathered}
\lambda_E\in\{0,0.1,\ldots,1\},\\
p_L\in\{0.1,0.25,0.5,0.75,1,1.5,2,3,4\}.
\end{gathered}
\label{eq:timing-search-grid}
\end{equation}
The parameters are evaluated by leave-one-session-out (LOSO)
cross-validation over the 31 formal sessions: in each fold, the pair
minimizing event-level MAE on the other 30 sessions is applied, without
refitting, to the held-out session, so that session's ratings are not used
to select the parameters applied to it.

\subsection{Results}
\label{app:timing-results}

\begin{figure}[tb]
\centering
\includegraphics[width=\linewidth]{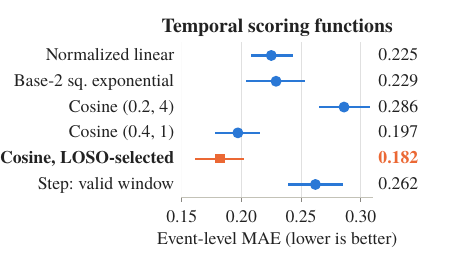}
\caption{\textbf{Agreement of temporal scores with human ratings.}
Event-level MAE with 95\% session-cluster bootstrap confidence intervals;
lower is better. Cosine settings are $(\lambda_E,p_L)$. The selected
cosine uses LOSO out-of-fold predictions.}
\label{fig:timing-function-results}
\end{figure}

The LOSO-selected cosine has the lowest MAE among the compared functions
(Figure~\ref{fig:timing-function-results}). Thirty of the 31 folds
selected $(\lambda_E,p_L)=(0.7,0.5)$ and one selected $(0.6,0.5)$. The
parameters used by Timely Correctness, $(0.7,0.5)$, are those selected on
all 31 formal sessions; the MAE reported for the selected cosine is its
out-of-fold value.

The selected cosine achieves MAE $0.182$, compared with $0.225$ for linear
scoring and $0.262$ for constant valid-window credit. Its paired difference
from linear is $-0.043$ with a 95\% confidence interval (CI) of
$[-0.064,-0.023]$, its improvement over
constant credit is $0.080$ ($[0.069,0.091]$), and it also improves over the
$(0.4,1)$ cosine setting by $0.015$ ($[0.005,0.025]$). These comparisons
support graded credit within the window. The selected $p_L=0.5$ lies
inside the search grid rather than at its lower boundary, and it gives a
shallower decline through moderate delay than $p_L=4$ while still
assigning zero at LU. Linear and selected cosine scores show similar
within-event ranking agreement ($\tau_b=0.740$ and $0.738$), but the
selected cosine achieves better calibration to the human rating scale, as
reflected by its lower MAE.

\paragraph{Spearman Correlation.}
The comparison with binary scoring in Section~\ref{sec:eval-timing} uses
Spearman's rank correlation computed within each event between a
function's scores and the mean human ratings of its six candidate moments,
averaged over events. Linear and selected cosine scores rank the six
moments identically in 78 of the 79 events and reach mean correlations of
0.833 and 0.832. Constant valid-window credit is identical across all six
moments of 12 events, where the correlation is undefined; over the
remaining 67 events it averages 0.645 (0.547 when undefined events count
as zero), and the fixed $[-6,+3]$\,s step averages 0.616 over its 50
defined events. Pooling all 474 moments instead gives 0.787 for both
graded scores and 0.586 for constant valid-window credit.

Selection using only the 46 necessary events from 27 sessions gives the
same parameters in all 27 LOSO folds, with out-of-fold MAE $0.178$ versus
$0.219$ for linear scoring, so including the optional events does not
change the selected setting. We use one shared function rather than
separate scoring functions for individual instructor categories, whose
small groups are partly confounded with the reactive/proactive
composition.

\subsection{Calibration and Rater Sensitivity Analyses}
\label{app:timing-rater-sensitivity}

The selected cosine keeps its advantage after adjusting for rater
differences and when any one rater is omitted. For the
calibration-informed analysis, we jointly modeled the valid calibration
and formal ratings with a random intercept for candidate moment and fixed
effects for study phase, rater, timing region (before the preferred
time, within the window, or after LU), intervention side, and the
corresponding rater interactions; the calibration ratings serve here only to estimate
rater severity and timing tolerance. The adjusted formal reference
correlates $0.993$ with the raw rater mean. Against it, cosine MAE is
$0.190$ and linear MAE is $0.236$, a paired difference of $-0.046$ (95\% CI
$[-0.063,-0.029]$). This agrees with the primary analysis. Given the small number of raters
and calibration events, we treat this adjustment as an exploratory
sensitivity analysis rather than independent validation.

The two-way mixed-effects absolute-agreement intraclass correlation
coefficient (ICC) is $0.663$ for one rater
(ICC(A,1)) and $0.855$ for the three-rater mean (ICC(A,3)). Omitting each
rater in turn preserves the cosine advantage, with MAE improvements of
$0.031$--$0.058$. Raters differed in strictness and preferred timing, which
supports using their mean rather than a single rater's curve.

\end{document}